\documentclass[lettersize,journal,fontsize=10]{IEEEtran}
\usepackage{amsmath,amsfonts}
\usepackage[linesnumbered,ruled,vlined]{algorithm2e}
\usepackage{tabularx} 
\usepackage{multirow} 
\usepackage{rotate}   

\usepackage{array}
\usepackage[caption=false,font=normalsize,labelfont=sf,textfont=sf]{subfig}
\usepackage{textcomp}
\usepackage{stfloats}
\usepackage{xcolor}
\usepackage{url}
\usepackage{verbatim}
\usepackage{graphicx}
\usepackage{hyperref}

\usepackage{caption}

\DeclareCaptionLabelFormat{lc}{\MakeLowercase{#1}~#2}
\newtheorem{definition}{Definition}

\newtheorem{remark}{Remark}
\newtheorem{proposition}{Proposition}

\def\BibTeX{{\rm B\kern-.05em{\sc i\kern-.025em b}\kern-.08em
    T\kern-.1667em\lower.7ex\hbox{E}\kern-.125emX}}
\usepackage{balance}
\begin{document}
\title{Partial Transport for Point-Cloud Registration}
\author{Yikun Bai\IEEEauthorrefmark{1}, Huy Tran\IEEEauthorrefmark{1}, Steven B. Damelin\IEEEauthorrefmark{2}, Soheil Kolouri\IEEEauthorrefmark{1}, Senior Member IEEE\\
  \IEEEauthorblockA{\IEEEauthorrefmark{1} Department of Computer Science, Vanderbilt University, Nashville, TN}\\
    \IEEEauthorblockA{\IEEEauthorrefmark{2} Michigan, Ann Arbor, MI}\\
}

\newcolumntype{C}{>{\centering\arraybackslash}X} 
\newcolumntype{Y}{>{\centering\arraybackslash}p{0.4cm}}
\maketitle

\begin{abstract}

Point cloud registration plays a crucial role in various fields, including robotics, computer graphics, and medical imaging. This process involves determining spatial relationships between different sets of points, typically within a 3D space. In real-world scenarios, complexities arise from non-rigid movements and partial visibility, such as occlusions or sensor noise, making non-rigid registration a challenging problem. Classic non-rigid registration methods are often computationally demanding, suffer from unstable performance, and, importantly, have limited theoretical guarantees. The optimal transport problem and its unbalanced variations (e.g., the optimal partial transport problem) have emerged as powerful tools for point-cloud registration, establishing a strong benchmark in this field. These methods view point clouds as empirical measures and provide a mathematically rigorous way to quantify the `correspondence' between (the transformed) source and target points. In this paper, we approach the point-cloud registration problem through the lens of optimal transport theory and first propose a comprehensive set of non-rigid registration methods based on the optimal partial transportation problem. Subsequently, leveraging the emerging work on efficient solutions to the one-dimensional optimal partial transport problem, we extend our proposed algorithms via slicing to gain significant computational efficiency, resulting in fast and robust non-rigid registration algorithms. We demonstrate the effectiveness of our proposed methods and compare them against baselines on various 3D and 2D non-rigid registration problems where the source and target point clouds are corrupted by random noise.

\end{abstract}


\section{Introduction}
Data acquisition and the increasing interest in augmented and virtual reality have led to an explosion of volumetric data, as point clouds exemplify. Point cloud data is prevalent in numerous applications, including robotics \cite{aguilar20173d,tache2009magnebike,pomerleau2015review}, autonomous driving \cite{ilci2020high,pan2019clustermap,poto2017laser}, medical imaging \cite{sinko20183d,rasoulian2012group} and computer graphics \cite{diez2012hierarchical,yuan20163d}. In many of these applications, the captured point clouds correspond to noisy observations of an object/scene undergoing different deformations. One of the core challenges in these applications is to perform point cloud registration, which refers to finding a transformation that aligns or partially aligns the source and target point sets. At a high level, any point cloud registration algorithm must solve two problems concurrently: 1) finding accurate correspondences between the points in the source and target point clouds  (implicitly or explicitly), and 2) modeling the deformation to match the corresponding source and target points. The existing methods then propose different correspondence estimation algorithms \cite{haehnel2003extension,domokos2011nonlinear,tsin2004correlation,li2008global,eisenberger2020deep,feydy2017optimal,dang2020learning} and/or propose novel deformation modeling \cite{besl1992method,yuan2020deepgmr,myronenko2010point,chui2003new,wang2022partial}.

The registration/deformation map (i.e., the transformation) could be rigid \cite{horn1987closed,horn1988closed,chen1992object,tsin2004correlation,huang2021comprehensive}, only involving translation and rotation, or non-rigid (e.g., an affine transformation or other nonlinear deformation) \cite{chui2003new,deng2022survey,jian2005robust,ma2015non}. Most existing works in the literature have focused on the rigid registration of point clouds, as it is a more prevalent problem in classic computer vision tasks such as  Simultaneous Localization And Mapping (SLAM) \cite{mendes2016icp,alismail2014continuous}. The core innovations in these approaches are often concerned with finding the right correspondences between the points. For instance, the classic Iterative Closest Point (ICP) algorithm \cite{horn1987closed,horn1988closed,chen1992object,zinsser2005point,du2007extension} relies on nearest-neighbor correspondences as measured via the Euclidean distance between points. However, the Euclidean distance between points might not be a reliable measure of proximity if, for instance, the point clouds are not approximately aligned or the point clouds are noisy. To establish better correspondences, many have looked into improving the similarity measures by defining features/descriptors for point cloud data\cite{deng20193d,deng2018ppfnet,qi2017pointnet++,zeng20173dmatch,yang2022learning,yew2020rpm,pauly2003multi}
that not only capture the location of a point but also encode the local geometry of the object in the vicinity of that point (e.g., curvature). These methods rely on nearest neighbor matching in an explicit or implicit feature space instead of the raw input space and are closely related to kernel methods \cite{tsin2004correlation,ma2015non,jian2005robust,myronenko2010point}. Moreover, to obtain robustness and avoid false correspondences, these methods often use the Random Sample Consensus (RANSAC) algorithm \cite{derpanis2010overview,fischler1981random} or its variations \cite{torr2000mlesac,chum2005matching}. Unfortunately, however,  RANSAC significantly increases the computational cost of the registration algorithm as it requires running the correspondence problem multiple times for different random subsets of the data. 

In many real-world applications, however, the deformation between two sets of point cloud data is inherently non-rigid. For instance, in medical imaging, the point cloud data could come from the surface of a tissue, which can undergo large non-rigid deformations. Such nonlinear deformations are generally modeled using two categories of approaches, parametric \cite{chui2003new,yang2011thin,tsin2004correlation} and non-parametric approaches \cite{myronenko2010point,myronenko2006non,wang2022partial}. In the parametric approaches, the deformation is characterized via a parametric function, e.g., parameters of an affine transformation \cite{tsin2004correlation} or Thin Plate Spline (TPS) parameters
 \cite{chui2003new}, and they are optimized to minimize the expected distance between the corresponding source and target points.  On the other hand, the non-parametric approaches directly calculate the displacement (velocity) between source points and their corresponding target points. The existing approaches for non-parametric non-rigid point cloud registration vary in how to estimate the velocity of each point and how to regularize the velocity vector field for coherency and smoothness\cite{myronenko2010point,yuille1988computational}. In this paper, we consider parametric non-rigid registration of point clouds and utilize Optimal Partial Transport (OPT) \cite{caffarelli2010free,figalli2010new}, an instance of the unbalanced optimal transport \cite{chizat2018unbalanced,fatras2021unbalanced}, as a unifying framework to achieve this task.


Solving the correspondence between points in the source and target point clouds is closely related to the celebrated optimal transport (OT) problem \cite{villani2021topics,Villani2009Optimal,peyre2019computational}. In short, treating point clouds as empirical distributions and given a transportation cost, the OT problem seeks the optimal assignment between the samples to minimize the expected transportation cost between the assigned samples. 
This principle has led to many OT-based point-cloud registration algorithms. However, a major limitation of OT is the mass preservation assumption, which requires all points in the source to be matched to all the points in the target point cloud. The mass preservation assumption limits the application of OT to problems where the points must be partially matched, e.g., noisy or occluded point clouds and, in general, partial registration problems \cite{dang2020learning}. Various ideas have been recently developed to allow the application of OT for partial registration \cite{wang2022partial,qin2022rigid}. For instance, some dynamically infer the mass of each particle (i.e., the importance of a point in the point cloud) \cite{shen2021accurate} so that the OT problem can ignore the particles with zero/small mass. Others have looked into defining outlier bins and solving the OT problem while allowing for matching points to the outlier bin \cite{dang2020learning}. This latter idea is deeply rooted in the unbalanced optimal transport problem, which allows for the creation and destruction of mass. In this paper, we use OPT and its sliced variation as an excellent match for partial and robust registration problems and demonstrate their performance in non-rigid partial registration of point clouds.




\noindent{\bf Contributions:} Our specific contributions are as follows. 
\begin{enumerate}
    \item A robust and unifying framework for parametric non-rigid registration using OPT. 
     \item Proposing a sliced-OPT framework for accelerated non-rigid registration between large-scale point clouds. 
     \item Demonstrating the performance of the proposed framework on various noisy point cloud registration problems and benchmarking them against baselines. 
\end{enumerate}


In what follows, we first review the related work in the literature in Section \ref{sec:related_work}. Next, in section \ref{sec:background}, we introduce some basic background of optimal transport and optimal partial transport. Following this section, we introduce our OPT-based methods in section \ref{sec:method} and demonstrate our methods in 2D and 3D experiments in \ref{sec:experiments}. Finally, we discuss the limitations and future direction in section \ref{sec:summary}. In appendix \ref{sec:notations}, we introduce all the notations in section \ref{sec:notations}.

\section{Related Work}
\label{sec:related_work}

The prior work on point cloud registration could be categorized based on the type of registration, e.g., rigid versus non-rigid, or based on how correspondences are calculated/found, e.g., feature-based approaches. A large body of existing work takes an alternating optimization approach, where they jointly estimate the correspondences and the transformation iteratively . These methods update the correspondences assuming the transformation is fixed and then update the transformation assuming that correspondences are fixed.

Alternatively, the recent and emerging work from the deep learning community on point cloud registration focus on 1) feature learning for calculating robust correspondences \cite{lecun2015deep,qi2017pointnet}, and 2) end-to-end estimation of the transformation \cite{liu2019flownet3d}.

Our proposed approach in this paper is to directly optimize the transformation by minimizing an optimal partial transportation metric between the transformed source and the target point clouds. Hence, it is closely related to the alternating optimization approaches, with the main difference that the calculation of correspondences is implicitly done when calculating the OPT metric. Below we briefly overview the rigid and non-rigid registration approaches, the emerging deep learning-based registration methods, and lastly, the existing OT-based registration algorithms. Finally, we differentiate our approach from the existing work and state our contributions. 
\vspace{-1em}
\subsection{Rigid/Affine Point Cloud Registration}
In rigid registration, one seeks a transformation composed of only a rotation and a translation to match the source and target point clouds. The dominant framework on this front is the seminal works of \cite{besl1992method} and \cite{zhang1994iterative} on Iterative Closet Point (ICP), and their numerous extensions \cite{rusinkiewicz2001efficient,zhang2021fast}. For a given initial transformation and a distance between the source and the target points (e.g., the Euclidean distance), the classic ICP algorithm finds the correspondences via the nearest neighbor assignment, and then it updates the transformation to be the least-squares minimizer. Given the correspondences, and using the Euclidean distance, finding the rotation and translation parameters has a closed-form solution. Through these alternating schema, ICP is guaranteed to converge to a locally optimal transformation. The classic ICP algorithm, see for example \cite{horn1987closed,horn1988closed,chen1992object,zinsser2005point,du2007extension} relies on a good initial transformation that provides nearly aligned point clouds so that the correspondences calculated with the nearest neighbor assignments are reliable. Moreover, the classic ICP algorithm fails in the presence of noise and when dealing with partial registration (i.e., when only a subset of points should be matched). Lastly, the classic ICP algorithm has a linear convergence rate, which can be too slow for various registration applications (e.g., for simultaneous localization and mapping (SLAM) in robotics).

To address the limitations of ICP, a large body of work has been developed in the past two decades. A critical idea in these works is to replace the binary assignment in ICP with relaxed soft assignments and provide a generalized version of the classic ICP algorithm. For instance, the Robust Point Matching (RPM) algorithm \cite{rangarajan1997robust,gold1998new} and its extensions. The footprints of the idea of using soft assignments can also be seen in numerous approaches that pose the registration problem as a \emph{maximum likelihood} estimation problem \cite{joshi1995problem,cross1998graph,luo2003unified,mcneill2006probabilistic}. An interesting connection to the OT theory here is that the entropy regularized Kantorovich problem \cite{cuturi2013sinkhorn}, also provides a soft assignment (i.e., the entropy regularized transport plan) that can be utilized for the iterative matching problem in registration. This idea is referred to as the Wasserstein Procrustes problem \cite{grave2019unsupervised,ramirez2020novel,jin2021two}. For instance, Grave et al. \cite{grave2019unsupervised} use this idea to align two high-dimensional point clouds for Natural Language Processing (NLP) applications. Another common idea for robustifying the \emph{maximum likelihood} based registration algorithms against noise, is to introduce an extra distribution term to account for noise/outliers \cite{rangarajan1997robust} (i.e., using dummy outlier bins).  An interesting connection to the OT theory, here, is the idea of introducing dummy source and sink points in the optimal partial transport theory (for instance, see \cite{chapel2020partial}) to allow for destruction and creation of mass in the source and target distributions, respectively. These connections motivated our approach to use OPT and sliced OPT for noisy point cloud registration. 
\vspace{-1em}
\subsection{Nonrigid registration}
Non-rigid registration techniques align two shapes that undergo non-rigid deformation, such as bending, stretching, and twisting. 
Unlike rigid registration, which only involves a single rotation and translation, non-rigid registration must often determine a deformation field \cite{deng2022survey}.
A proper representation of the deformation fields must be chosen to balance sufficient expressiveness for accurate alignment and computational efficiency. Generally, the deformation fields can be divided into two categories: non-parametric and parametric models. 
In the following, we introduce some widely used representation techniques for deformation. 

 \textbf{Cohrent point drift (CPD) method}.
 The coherent point drift (CPD) method is proposed by \cite{myronenko2006non, myronenko2010point}. In this method, the point clouds are regarded as Gaussian mixture models (GMMs), and they have aligned via minimizing the discrepancies between them\footnote{In different literature, CPD is attributed into RKHS-based method, non-parametric approach, or probabilistic (GMM) model-based method. Here, to avoid controversial classifications, we treat it as a separate category.}. In this work, the deformation can be formulated as  
$$\hat{y}=x+\sum_{k\in \mathcal{I}}\alpha^k\phi_G(x,x^k),$$
where $\mathcal{I}\subset[1:n]$ is a subset of (index of) source points $\phi_G$ is the Gaussian kernel, which is introduced in Appendix \ref{sec:notations}.  
 Hirose et al. \cite{hirose2020bayesian} proposed a variant of the CPD method in a Bayesian setting. The new method is proved to have a convergence guarantee under variational Bayesian inference. Wang et al. \cite{wang2022partial} use the Gaussian kernel matrix in the CPD method to regularize the displacement parameters, and the correspondence is estimated by optimal partial transportation. 
 
In addition, deep learning techniques have been applied in this framework. For example, 
\cite{wang2019coherent}, proposes the so-called CPD-network method, which can learn a displacement field function to estimate a certain geometric transformation from a training dataset. Thus, these techniques can predict the desired geometric transformation to align previously unseen pairs of data without an additional optimization process. 
Similarly, \cite{ma2018nonrigid} combines CPD with a certain transformer network technique in this method.



\textbf{Thin plate spline-based methods.} 
This type of method uses a deformation field defined by the so-called thin plate spline function \cite{bookstein1989principal,wahba1990spline,duchon1977splines} in $\mathbb{R}^3$ space. In particular, 
$$f(x)=\sum_{k=1}^K \alpha^k\phi_T(x^k,x)+B^Tx+\beta,$$
where $\phi_T$ is TPS kernel, $B\in\mathbb{R}^{D\times D},\beta\in\mathbb{R}^D$, we refer appendix \ref{sec:notations} and subsection B in section \ref{sec:method} for details. 
The above formulation is derived from a square error minimization problem with a second derivative regularization term.  

In particular,  \cite{chui2003new} develops the popular TPS-RPM (thin plate spline-robust point-matching) algorithm which uses the TPS as the non-rigid spatial mapping and the soft assign for the correspondence. \cite{domokos2011nonlinear} traces back the registration problem to the solution of a system of nonlinear equations
that directly gives the parameters of the TPS function. Thus the proposed method recovers deformation without established correspondence. In addition, \cite{huang2019automatic} developed a new technique that chooses the control points based on the feature correspondence between two surfaces. 

\textbf{Kernel correlation and probability model}.  This type of method employs kernel correlation function to ascertain correspondences between point clouds or features. Typically, these point clouds are often represented using probability models. The seminal work proposed by \cite{tsin2004correlation} set the stage for subsequent advancements in this type of approach. This study leverages the kernel correlation to the similarity between the source and target point set and formalizes the registration challenge as a kernel correlation optimization issue. Building on this foundation, \cite{jian2005robust} employed the Gaussian Mixture Model (GMM) to characterize each data point cloud. Their approach to the registration problem aimed to minimize the discrepancy, especially the L2 distance, between the two GMMs. Moreover, they integrated Thin Plate Splines (TPS) and Gaussian radial basis functions for non-rigid transformation models. The CPD method, cited in \cite{myronenko2010point,ma2015non}, is a subsequent development. This method adopts the GMM model to represent the point cloud and determines correspondences by solving a maximum likelihood problem. In recent times, point cloud kernel correlation has been merged with deep learning models. For instance, \cite{shen2018mining} enhances the \textit{Pointnet} - a stage-of-art neural network model for semantic learning on 3D point clouds by integrating a kernel correlation layer, allowing the model to harness the local geometric structures of point clouds.

\textbf{ICP-FFD}
The ICP algorithm is widely used for rigid registration but has numerous limitations, as discussed earlier. To extend its capabilities, various techniques have been developed to model local deformations. Abdelmunim and Farag \cite{abdelmunim2011elastic} provide a flexible non-rigid registration method by combining ICP with Freeform deformation (FFD). FFD defines a lattice of control points, which can be moved to control the deformation of the underlying point cloud. The ICP-FFD  method iteratively refines the alignment between point clouds by minimizing the distance between corresponding points while adjusting the control points' positions to account for non-rigid transformations. In practice, this approach has been applied widely, including in medical image registration and facial expression recognition. Another combination is ICP and Gaussian Process Regression (GPR) \cite{brown2007global}. GPR is a powerpoint-cloud learning technique that can model the non-rigid deformation between two point clouds as a smooth, non-linear function. Like ICP-FFD, the ICP-GPR algorithm also iteratively refines the alignment between the point clouds using ICP while modeling the non-rigid deformation with GPR. This approach can handle large deformations while providing an explicit representation of the deformation function, which helps with interpretability. ICP can also be combined with an energy function to handle non-rigid registration \cite{li2008global}. 

The ICP algorithm is then used iteratively to minimize the energy function, resulting in a registration that considers both the point cloud alignment and deformation smoothness.

\textbf{Large deformations}.
In non-rigid registration, large deformations refer to a class of transformations that may change the overall structure of the point cloud significantly. Isometric deformations are a specific type of large deformations where the intrinsic geometric properties, such as geodesic distances between points, are preserved, whereas the extrinsic properties, like Euclidean distances, may change. For instance, in their seminal work, Adams et al. \cite{huang2008non} proposed a large deformation framework based on the theory of functional maps. Their proposed method establishes correspondences between two shapes by aligning their Laplace-Beltrami eigenfunction bases, which are intrinsic geometric operators invariant under isometric deformations. The method consists of several key steps, including computing the Laplace-Beltrami operators, obtaining eigenbases for representing shapes, representing shapes as functions on their respective surfaces, finding a transformation, and establishing point correspondence. The method has been extended in various ways with additional geometric or topological information to handle complex deformations \cite{sharma2011topologically,alhashim2015deformation}. It has been successfully applied widely in practice for registration, object recognition\cite{cho2014finding}, and computer animation \cite{levi2014smooth,tevs2012animation}.

\vspace{-1em}
\subsection{Deep Learning Based Registration}

Deep learning has rapidly emerged in recent years as a powerful tool for solving many computer vision tasks, including point cloud registration \cite{zhang2020deep}. Recent advances in the field have led to a plethora of neural network-based approaches that offer improved speed \cite{qi2017pointnet}, accuracy \cite{wang2019non}, and robustness \cite{aoki2019pointnetlk} for registration. In fact, these methods demonstrate great promise in enhancing the pipeline and overcoming challenging scenarios where traditional methods fall short. 

Existing approaches often involve finding correspondences in 3 key steps: feature selection, matching, and motion estimation \cite{besl1992method} \cite{rusu20113d}. 
First, for \textbf{feature selection}, the traditional methods have evolved from using basic Cartesian coordinates \cite{besl1992method} \cite{zhang1994iterative} to hand-crafted descriptors capturing complex geometric properties \cite{johnson1999registration} \cite{hoppe1992surface} \cite{pauly2003multi}. These traditional methods are often sub-optimal for large datasets. In sharp contrast, deep learning approaches like PointNet \cite{qi2017pointnet} and its extensions \cite{qi2017pointnet++} offer automated, robust feature extraction by leveraging neural networks \cite{lecun2015deep}. Regarding the second step, \textbf{matching},  traditional methods, like ICP, struggle with issues such as sensitivity to initialization, convergence to local minima, and poor performance in partial matching. while Random Sample Consensus (RANSAC) requires balancing accuracy and computational cost. Deep learning-based solutions like Deep Closest Point (DCP) \cite{wang2019deep} offer robust and efficient matching by leveraging PointNet-based architectures and attention mechanisms.
For the third step, \textbf{motion estimation}, traditional methods for motion estimation, such as least square linear regression and Singular Value Decomposition (SVD), are sensitive to noise and limited to rigid transformations. In contrast, deep learning-based solutions like FlowNet3D \cite{liu2019flownet3d} offer robust and flexible motion estimation by learning point-wise correspondence flow between point clouds.

\textbf{Correspondence-free methods}
There are also deep learning approaches that completely bypass finding correspondences by regressing motion parameters using global features. For example, \textit{PointNetLK} \cite{aoki2019pointnetlk} and \textit{PCRNet} \cite{sarode2019pcrnet} both utilize deep learning techniques to effectively estimate the relative pose between two point clouds without explicitly finding point correspondences. \textit{PointNetLK} \cite{aoki2019pointnetlk} extends the original \textit{PointNet} \cite{qi2017pointnet} architecture by incorporating a differentiable Lucas-Kanade (LK) layer to estimate the transformation parameters iteratively. \textit{PCRNet} \cite{sarode2019pcrnet}, on the other hand, employs a Siamese network architecture to learn global point cloud features and regress the transformation parameters directly. \textit{PCRNet} leverages the robustness of \textit{PointNet} for feature extraction and incorporates a fully connected regression network to predict the 6-DoF transformation. Some other methods are completely end-to-end, such as \textit{DeepVCP} \cite{lu2019deepvcp}, allowing prediction of the correspondences on raw data, without pre- or post-processing.

Despite their effectiveness, deep learning-based supervised approaches have several significant limitations. For instance, they require a large amount of annotated training data, which could be prohibitively expensive with scale. Our work in this paper is orthogonal to the research in the deep learning community, as our method is also compatible with these approaches, and can be integrated into learning-based pipelines.  


\vspace{-0.5em}
\subsection{OT Based Point Set Registration}

Optimal transport (OT) is a mathematical approach that can compare two probability measures. In the context of point clouds, they can be treated as discrete probability measures, and the OT problem can be utilized to determine the optimal (soft) correspondences between two sets of points. These correspondences aim to minimize the expected transportation cost, which is the cost of moving from one point to another. After obtaining the optimal correspondences between the points in the two point clouds, it becomes possible to compute the transformation that aligns the two point clouds \cite{feydy2017optimal}. This transformation can be rigid, such as translation, rotation, scaling, or non-rigid, like general deformations.

For example: \cite{puy2020flot} proposed a method to estimate the scene flow on point clouds using optimal transportation and deep features. On synthetic and real-world datasets, they demonstrate that the new method can perform as well as the state-of-art existing methods while requiring much fewer parameters and without multi-scale analysis. \cite{mei2021cotreg} proposes a learning framework for predicting correspondences of 3D point cloud registration by transferring point-wise matching and structural matching into a Wasserstein distance-based and a Gromov-Wasserstein distance-based optimizations, respectively. The proposed framework can accurately predict correspondences of 3D point cloud registration and achieve state-of-the-art performance on several benchmarks. In \cite{eisenberger2020deep}, they use entropic regularized optimal transportation and the smooth shells method to estimate the unsupervised 3D shape correspondence. 

In addition to classical OT in the balanced setting, as mentioned above, unbalanced OT has been widely applied in shape registration problems too. 
\cite{shen2021accurate} investigates the use of Hellinger Kantorovich distance (the so-called robust optimal transport in the paper) for shape matching. The TPS-RPM (thin plate sprine-robust point-matching) algorithm proposed by \cite{chui2003new} essentially applied optimal partial transport distance to estimate the correspondence for the non-rigid registration problem. Similarly, \cite{qin2022rigid} utilize optimal partial transport via a hard marginal constraint for solving non-rigid problem. Furthermore, sliced-OPT based approaches have been proposed recently  \cite{Bonneel2019sliced} 
\cite{bai2022sliced}. 
 
\textbf{OT in Deep Learning:} Wang and Solomon (2019) proposed Deep Closest Point (DCP) \cite{wang2019deep}, a deep learning-based approach for point cloud registration that employed the Sinkhorn algorithm to minimize the regularized OT distance between two point sets. DCP utilized PointNet to learn local geometric features and incorporated a differentiable module based on the Sinkhorn algorithm for the alignment process. This end-to-end trainable architecture outperformed traditional ICP-based methods. Since the introduction of DCP, several follow-up works have emerged, addressing different challenges and further improving the performance of point cloud registration methods that combine OT and deep learning. Yuan et al. (2020) presented DeepGMR \cite{yuan2020deepgmr}, which used deep learning to compute a Gaussian Mixture Representation (GMR) of point clouds and applied OT to align the GMRs. This approach reduced computational complexity while maintaining high registration accuracy. Yew and Lee (2020) introduced RPM-Net \cite{yew2020rpm}, a method that combined deep learning and the Robust Point Matching (RPM) framework. RPM-Net employed a learned affinity matrix based on deep features and solved the registration problem using a differentiable OT-based module. These works demonstrate the potential of combining OT and deep learning for point cloud registration, leading to improved performance over traditional methods. However, further research is needed to address challenges such as computational complexity, partial overlap constraint, and lack of interpretability.

\vspace{-1em}
\section{Background}
\label{sec:background}

\subsection{Optimal Transport}
Optimal transport (OT) theory, pioneered by Monge \cite{monge1781memoire} and Kantorovich \cite{kantorovich1942translocation,kantorovich1948problem}, studies the most cost-efficient way to move mass from one measure to another measure. It has attracted abundant attention in data science, statistics, machine learning,
signal processing and computer vision. 

\subsubsection{Classic Optimal Transport}
The classic optimal transport, known as \textbf{Kantorovich formulation}, is defined as follows:  let $\mathcal{P}(\mathbb{R}^D)$ denote the set of all Borel probability measures defined in $\mathbb{R}^D$. Given two probability measures, $\mu,\nu\in\mathcal{P}(\mathbb{R}^D)$, and $c:(\mathbb{R}^D)^2\to \mathbb{R}_+$ a lower semi-continuous function denoting the transportation cost, the optimal transport problem is defined as: 
\begin{align}
    \text{OT}(\mu,\nu):=\inf_{\gamma\in\Pi(\mu,\nu)} c(\hat{y},y)d\gamma(\hat{y},y) \label{eq: OT}
\end{align}
where $\Pi(\mu,\nu)\subset \mathcal{P}((\mathbb{R}^D)^2)$ is the set of joint distribution whose marginals are $\mu,\nu$, respectively. We mathematically denote the two marginals of $\gamma$ as
$(\pi_1)_\#\gamma =\mu, (\pi_2)_\#\gamma = \nu$, where $\pi_1$, $\pi_2$ are canonical projection maps, and for any (measurable) function $f: (\mathbb{R}^D)^2\to \mathbb{R}^D$,
$f_\#\gamma$ is the push-forward measure defined as 
$f_\#\gamma(A)=\gamma(f^{-1}(A))$ for any Borel set $A\subseteq\mathbb{R}^D$.

Regarding the point cloud registration scenario, we use $\mu,\nu$ to represent the transformed source point cloud and target point cloud, that is, we set 
$\mu,\nu$ as empirical distributions: 
$\mu=\frac{1}{N}\sum_{n=1}^N\delta_{\hat{y}^n},\nu=\frac{1}{M}\sum_{m=1}^M\delta_{y^m}$, where $\delta_{\hat{y}}$ is the Dirac delta function, $\{\hat{y}^n\in \mathbb{R}^D\}_{n=1}^N$ and $\{y^m\in \mathbb{R}^D\}_{m=1}^M$ are distinct points, the above formulation becomes: 
\begin{align}
OT(\mu,\nu):=\inf_{\gamma\in\Gamma(\frac{1}{N}1_{N},\frac{1}{M}1_M)}\sum_{n=1}^N\sum_{m=1}^M c(\hat{y}^n,y^m) \gamma_{n,m} \label{eq: empirical OT}
\end{align}
where $c(\hat{y}^n,y^m)$ denote the distance between points $(\hat{y}^n,y^m)$, $\gamma$ defines an ``soft'' correspondence between the two point clouds. 

By OT's theory, the minimizer for problem \eqref{eq: OT} (and \eqref{eq: empirical OT}) always exists, and we can replace $\inf$ by $\min$. Intuitively, for every feasible joint distribution $\gamma$ in problem \eqref{eq: empirical OT} (i.e. $\gamma$ satisfies the constraints in the problem), $\text{supp}(\gamma)$ defines a correspondence between $\mu,\nu$ and $\gamma$ describes a transportation plan from $\{\hat{y}^n\}_{n=1}^N$ to $\{y^m\}_{m=1}^M$. Indeed, for each $(n,m)\in[1:N]\times [1:M]$, the value of $\gamma_{n,m}$ denotes the amount of mass that will be moved from $\hat{y}^n$ to $y^m$ and thus $\gamma_{n,m}=0$ denotes that there is no correspondence between $\hat{y}^n$ and $y^m$. 
When a transportation plan transport masses from $\hat{y}^n$ to at least two $y$ points, say $y^{m_1}, y^{m_2}$, (i.e.,  $\gamma_{n,m_1},\gamma_{n,m_2}>0$), we say there is mass splitting in this transportation plan $\gamma$ or the correspondence between $\mu$ and $\nu$ is soft. Mass splitting is allowed in Kantorovich's formulation of the OT problem; however, in some particular cases, for example, when $N=M$, there is no need to consider such a plan. In particular, when $M=N$ and all samples have uniform mass $1/N$, by OT's theory (see page 5 in \cite{villani2021topics}), the optimal $\gamma$ is an $N\times N$ permutation matrix (i.e., $\gamma_{n,m}\in\{0,1\},\forall (n,m)$ and for each row and column, there exists only one $\gamma_{n,m}=1$), which implies that a one-to-one mapping induces the OT plan. Therefore, the above Kantorovich's formulation is equivalent to the following so-called \textbf{Monge formulation}: 
\begin{align}
MOT(\mu,\nu):=\inf_{\substack{L:[1:n]\to [1:n]\\
    L \text{ is }1-1}}\sum_{n=1}^Nc(\hat{y}^n,y^{L(n)}))
\end{align}
and the optimal $L$ is called Monge's mapping. Regarding the registration scenario, $L$ defines a one-by-one correspondence between the source and target point clouds. 

When $N\neq M$, (for convenience, say $N<M$), feasible $\gamma$s would have mass splitting, and Monge's mapping does not exist.

In shape registration tasks, we still prefer hard correspondences; we can use barycentric projection to get an approximation of the Monge mapping. In particular, we find the optimal transportation plan $\gamma^*$ for $OT(\mu,\nu)$ and we call $\hat\nu:=\frac{1}{N}\sum_{n=1}^N \delta_{((M\gamma Y)[n,:])^T}$ (where $Y=[y^1,\ldots y^M]^T$) as the barycentric projection \cite{ambrosio2005gradient,bai2023linear} of $\nu$ with respect to $\mu$, which is an $N-$points empirical distribution and can be regarded as the ``representation'' of $\nu$. Furthermore, if $N=M$ (i.e. Monge's mapping exists), then 
$$OT(\mu,\nu)=\sum_{n=1}^N c(\hat{y}^n,(M\gamma Y)[n,:]^T),$$ where $\gamma Y\in\mathbb{R}^{N\times D}$ is matrix multiplication, $M$ is a scalar, and $(M\gamma Y)[n:]$ is $n^{th}$ row of matrix $M\gamma Y$. Thus, in practice, the Monge mapping between $\mu$ and $\hat\nu$ can be used to approximate the optimal transportation plan $\gamma$ for $OT(\mu,\nu)$.

\begin{figure}[t!]
    \centering    \includegraphics[width=\linewidth]{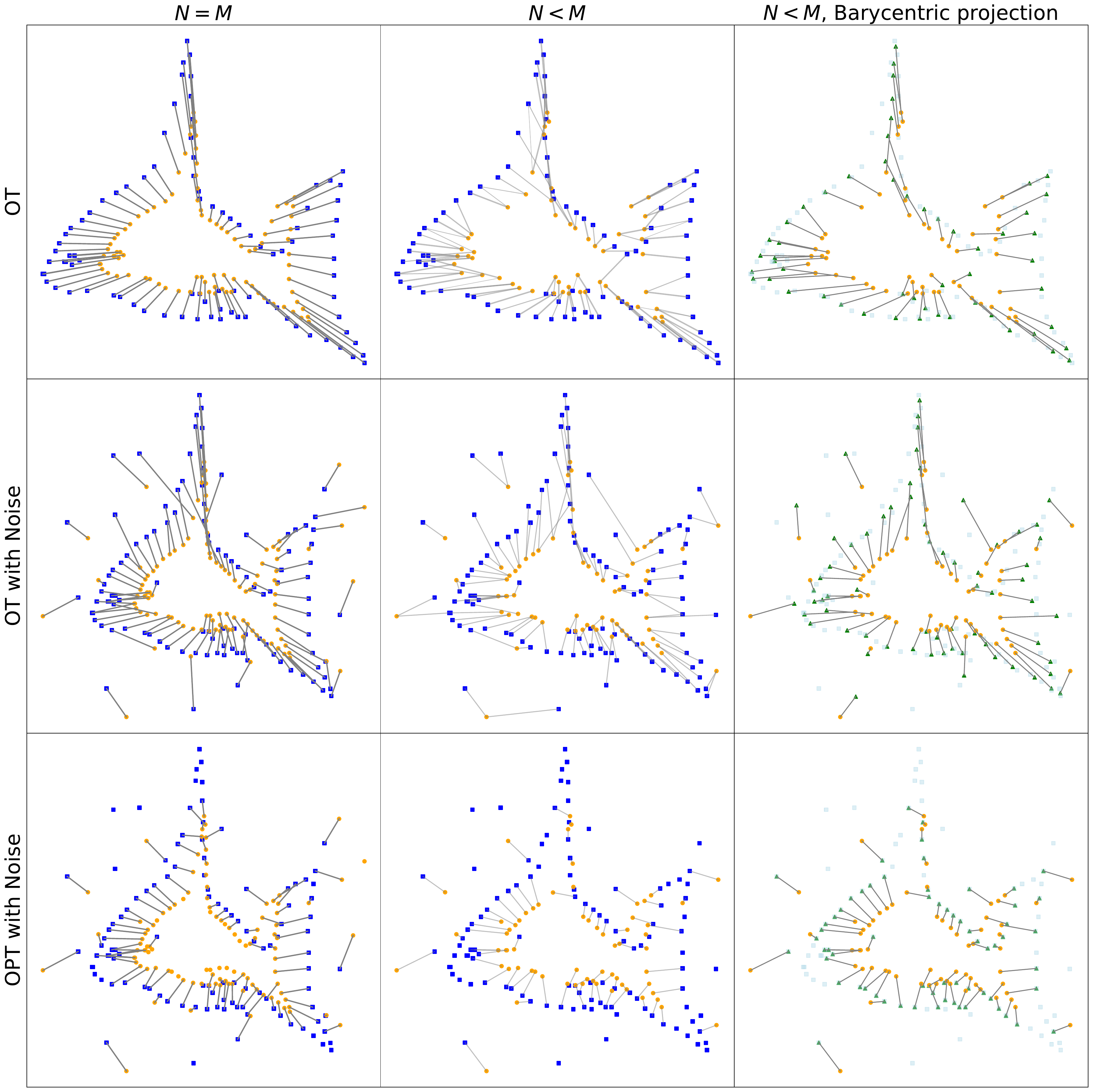}
    \caption{In the first column $(N=M)$, the optimal transportation plan provides a hard assignment where every point has at most one correspondence. In the second column $(N<M)$, the optimal transportation plan induces a soft assignment, which can be turned into a hard assignment with the barycentric projection of $\nu$, as shown in the third column.}
    \label{fig:my_label}
\end{figure}

\subsubsection{Optimal Partial Transport}
The OPT problem, in addition to mass transportation, allows mass destruction at the source and mass creation at the target. Here the mass destruction and creation penalty will be linear. Let $\mathcal{M}_+(\mathbb{R}^D)$ denote the set of all positive Radon measures defined on $\mathbb{R}^D$, suppose $\mu,\nu\in\mathcal{M}_+(\mathbb{R}^D)$, and $\lambda\ge 0$, the OPT problem is then defined as:
\begin{align}
\text{OPT}_{\lambda}(\mu,\nu):=\inf_{\gamma\in\Gamma_\leq(\mu,\nu)} \int c(\hat{y},y) \, d\gamma +\lambda(|\mu|+|\nu|-2|\gamma|), \label{eq: OPT}
\end{align}
 for $\Gamma_\leq(\mu,\nu):=\{\gamma\in \mathcal{M}_+((\mathbb{R}^D)^2) | \pi_{1\#}\gamma\leq \mu, \pi_{2\#}\gamma\leq \nu\}$ where $\pi_{1\#}\gamma \leq \mu$ denotes that for any Borel set $A\subseteq\mathbb{R}^D$, $\pi_{1\#}\gamma(A)\leq \mu(A)$, and we say $\pi_{1\#}\gamma$ is \textit{dominated by} $\mu$, analogously for $\pi_{2\#}\gamma\leq \nu $; the notation $|\mu|$ denotes the total mass of measure $\mu$.
 When $c(\hat{y},y)$ is a metric, $\text{OPT}_{\lambda}(\cdot,\cdot)$ is a metric on $\mathcal{M}_+(\mathbb{R}^D)$ (see~\cite[Proposition 2.10]{chizat2018unbalanced}, \cite[Proposition 5]{Piccoli2014Generalized}, \cite[Section 2.1]{lee2021generalized} and~\cite[Theorem 4]{chen2017matricial}). 

In point cloud registration problem, we set $\mu,\nu$ to be 
empirical measures,~$\mu=\sum_{n=1}^N\delta_{\hat y^n}$ and $\nu=\sum_{m=1}^M\delta_{y^m}$, with distinct $\hat y^n,y^m\in \mathbb{R}^D,\forall n,m$ to denote the estimated and target point clouds. 
The OPT problem \eqref{eq: OPT}, denoted as $\text{OPT}(\{\hat{y}^n\}_{n=1}^N,\{y^m\}_{m=1}^M)$, can be rewritten as:
\begin{align}
&\text{OPT}(\{\hat{y}^n\}_{n=1}^N,\{y^m\}_{m=1}^M)\nonumber\\
:=&\min_{\pi\in\Gamma_{\leq}(1_N,1_M)} \sum_{n,m} c(\hat{y}^n,y^m)\gamma_{n,m}+\lambda(N+M-2\sum_{n,m}\gamma_{n,m}) \label{eq: OPT empirical}
\end{align}
where
\[ \Gamma_{\leq}(1_N,1_M):= \{\gamma\in \mathbb{R}_+^{N\times M}: \gamma 1_M\leq 1_N, \gamma^T1_N\leq 1_M\}, \] 
and $1_N$ denotes the $N\times 1$ vector whose entries are $1$ and analogously for $1_M$.

It has been shown that the optimal plan $\gamma$ for the empirical OPT problem is induced by a 1-1 mapping \cite{bai2022sliced}. Thus, the above problem can be further simplified as follows: 
\begin{align}
\text{OPT}(\{\hat{y}^n\}_{n=1}^N,\{y^m\}_{m=1}^M)&=\min_{L}\sum_{n\in \text{Dom}(L)}c(\hat{y}^n,y^{L(n)})\nonumber\\ 
&+\lambda(N+M-2|\text{Dom}(L)|)
\end{align}
where $L: [1:N]\hookrightarrow [1: M]$ is a partial bijection, i.e. $\text{Dom}(L)\subset [1:N]$, $L$ is a 1-1 mapping. $|\text{Dom}(L)|=\#\text{Dom}(L)$ is the cardinality of the set \text{Dom}(L). Note in the registration problem, $L$ denotes a partial correspondence. That is, we only define the correspondence for points in the domain and range of $L$.

\subsubsection{Primal-form of OPT}
If we replace the penalty term of \eqref{eq: OPT} with a constraint, i.e.~we impose the condition $|\gamma|= \zeta$, \footnote{In the original formulation, this constraint is $|\gamma|\ge \zeta$. It is straightforward to verify the equivalence. Thus in this paper, we do not distinguish the conditions ``$\ge\zeta$'' and ``$=\zeta$''.}, where $\zeta\in[0, \min (|\mu|,|\nu|)]$ is a constant, then \eqref{eq: OPT} is closely related to the Lagrangian formulation of the following ``primal problem'': 
\begin{align}
\text{Primal-OPT}(\mu,\nu;\zeta)=\inf_{\gamma\in\Pi_\leq(\mu,\nu)}&\int c(\hat{y},y) d\gamma(\hat{y},y)  \label{eq: OPT primal} \\
&\text{s.t. } |\gamma|= \zeta. \nonumber
\end{align}

For empirical measures $\mu=\sum_{n=1}^N\delta_{\hat y^n}, \nu=\sum_{m=1}^M\delta_{y^m}$, the above problem can be written as 

\begin{align}
\text{Primal-OPT}(\mu,\nu;\zeta)=\min_{\gamma\in\Pi_\leq(1_N,1_M)}& \sum_{n,m}c(\hat{y}^n,y^m) \gamma_{n,m}  \label{eq: OPT primal empirical}\\ 
&\text{s.t. } |\gamma|= \zeta. \nonumber  \end{align}

\subsubsection{Solvers and their computational complexities}
For empirical OT \eqref{eq: empirical OT}, its objective function is linear, and the feasible space (for $\gamma$) is a polytope. Thus it can be solved by \textbf{Linear programming} \cite{karmarkar1984new}. Furthermore, as shown in \cite{caffarelli2010free}, OPT can be formulated as a balanced OT problem by introducing \textit{reservoir} points, thus empirical OPT \eqref{eq: OPT empirical} and empirical primal-OPT \eqref{eq: OPT primal empirical} \footnote{We refer the Appendix \ref{sec:opt} for the computation of primal-OPT.} could also be solved by linear programming. However, the time complexity of linear programming is $\mathcal{O}(NM(N+M))$.

One idea to facilitate linear programming is to convert the linear optimization problem into a strictly convex problem. The most successful approach in this area is \textbf{Entropy Regularization}, which adds the transport plan's entropy to the OT objective function and then applies the Sinkhorn-Knopp algorithm \cite{sinkhorn1964relationship,cuturi2013sinkhorn}. The algorithm can be extended to the large-scale stochastic \cite{genevay2016stochastic} and unbalanced settings \cite{benamou2015iterative,chizat2018scaling}.
For moderate regularization, these algorithms converge fast, however, there is a trade-off between accuracy versus stability and convergence speed for small regularization parameters. In practice, under an unsuitable regularization, the time complexity of the Sinkhorn algorithm could, unfortunately, be even worse than linear programming. 

Other notable approaches include restricting the cost function to specific metrics \cite{guittet2002extended,orlin1988faster,sato2020fast}, or enforcing low-rank constraints on $\gamma$ \cite{scetbon2021low,scetbon2022low}.  In particular, when the transportation cost is a metric, \textbf{network flow methods} \cite{guittet2002extended,orlin1988faster} can be applied, or when $c$ is a tree metric, an efficient algorithm based on dynamic programming with time complexity $\mathcal{O}(n\log^2n)$ is proposed in \cite{sato2020fast}. However, in high dimensions existence (and identification) of an appropriate metric tree remains challenging. Lastly, low-rank approximations \cite{scetbon2021low,scetbon2022low} are shown to lead to very fast solvers, however, at the cost of accuracy.

\subsubsection{Sliced Optimal (Partial) Transport}
When the space is ordered (e.g. 1 Dimensional Euclidean space) and the cost function is consistent with respect to the order, i.e. $c(x^1,x^2)\leq c(x^1,x^3)$ if $x^1\leq x^2\leq x^3$, then by OT's theory \cite{villani2021topics}, the balanced OT problem has a closed form solution, i.e., the increasing re-arrangement function given by the north-west corner rule. Utilizing this theory, \textbf{Sliced OT} techniques \cite{rabin2011wasserstein,kolouri2015radon,bonneel2015sliced,kolouri2016sliced,liutkus2019sliced} are proposed with the main idea to calculate the expected OT distance between 1-dimensional marginal distributions (i.e., slices) of two $d$-dimensional distributions. The expectation is numerically approximated via a Monte Carlo integration scheme. Other notable extensions of these distances include the generalized, the max-sliced, and the convolutional sliced Wasserstein distances \cite{kolouri2019generalized,deshpande2019max,nguyen2022revisiting}. 

Sliced OT techniques can be extended into the unbalanced transport setting \cite{bai2022sliced,sejourne2023unbalanced}. In particular, Bai et al. \cite{bai2022sliced} define the sliced optimal partial transport (SOPT) as follows: 
\begin{definition}
In $\mathbb{R}^D$ space, given $\mu,\nu\in\mathcal{M}_+(\mathbb{R}^D)$ and $\lambda:\mathbb{S}^{D-1}\to \mathbb{R}_{++}$ is an $L_p$ function, we define the sliced optimal partial transport (SOPT) problem as follows:
\begin{align}
\text{SOPT}_\lambda(\mu,\nu)=\int_{\mathbb{S}^{D-1}} \text{OPT}_{\lambda(\theta)}(\langle \theta,\cdot\rangle_\# \mu, \langle \theta,\cdot \rangle_\# \nu) d\sigma(\theta) \label{eq: sliced-opt}
\end{align}
where $\text{OPT}_\lambda(\cdot,\cdot)$ is defined in \eqref{eq: OPT}, 
$\sigma\in\mathcal{P}(\mathbb{S}^{D-1})$ is a probability measure such that $\text{supp}(\sigma)=\mathbb{S}^{D-1}$.   
\end{definition}
The 1D OPT problem $\text{OPT}_{\lambda(\theta)}(\langle \theta,\cdot\rangle_\# \mu,\langle\theta,\cdot\rangle_\# \nu)$ can be solved efficiently by the algorithm 1 in \cite{bai2022sliced}. Similar to OPT problem, when cost function $c$ is a ($p$-power of a) metric, SOPT \eqref{eq: sliced-opt} defines a metric in $\mathcal{M}_+(\mathbb{R}^D)$. 

\vspace{-1em}
\subsection{Procrustes Alignment Problem}

Procrustes analysis \cite{schonemann1966generalized,damelin2024near} is a well-known method to learn a linear transformation between two sets of matched points (point clouds) $X\in \mathbb R^{N\times D}$ and $Y\in \mathbb R^{N\times D}$.  If the correspondences between the two sets are known (i.e., which point
of $X$ corresponds to which point of $Y$), and for convenience, suppose the correspondence is induced by identity matrix (i.e. $x^n\mapsto y^n,\forall n$),
then this linear transformation can be recovered by solving
the least square problem: ${\rm min}_{W\in \mathbb R^{D\times D}}\|XW-Y\|_2^2$. The case 
${\rm min}_{W\in O_D}\|XW-Y\|_2^2$ where $O_D$ is the set of orthogonal matrices has a closed form solution discovered by Schönemann \cite{schonemann1966generalized} given by $W^*=UV^{T}$ where $USV^{T}$ is the singular value decomposition of $X^{T}Y$. The orthogonality ensures that the distances between points are unchanged by the transformation. This problem is called the ``Orthogonal Procrustes Problem.''

 Procrustes alignment problem assumes known correspondences between source and target points. When these correspondences are not known, the optimal transport (or optimal partial transport) can be used to first obtain correspondences (or soft correspondences) between the source and target and then solve for the linear transformation $W$. This problem is referred to as ``Wasserstein Procrustes'' \cite{grave2019unsupervised}. 
 Procrustes Wasserstein is formalized as:
 $${\rm min}_{W\in O_D}{\rm min}_{P\in {\cal P}_N}\|XW-PY\|_2^2$$ where ${\cal P}_N$ is the set of $N\times N$ permutation matrices. While for a fixed $P$ the problem is convex in $W$, and vice versa, the problem is, unfortunately, not jointly convex in $W$ and $P$. Hence,  stochastic alternating optimization techniques are utilized to solve this optimization problem \cite{grave2019unsupervised}. 
 The rigid registration problem between source point cloud $X$ and target point cloud $Y$ can be regarded as the above Wasserstein Procrustes problem, and the transformation model is defined as $x\mapsto W^Tx.$
 






\vspace{-1em}
\section{Our Method}\label{sec:method}
In space $\mathbb{R}^D$, we consider the registration map, $f: \mathbb{R}^D\to\mathbb{R}^D$, as follows: 
\begin{align}
f(x):=\sum_{k=1}^K\alpha^k\phi(x,c^k)+B^Tx+\beta \label{eq: model general}
\end{align}
where $\alpha^k\in\mathbb{R}^D$ for each $k$, $\phi: (\mathbb{R}^D)^2\to \mathbb{R}$ is a smooth kernel function \footnote{We refer Appendix \ref{sec:notations} for the introduction of kernel function. The TPS kernel will also be discussed in subsection B.}, $B\in\mathbb{R}^{D\times D}$ modeling the linear portion of the deformation, $\beta\in\mathbb{R}^D$ is the translation parameter, $c^1,..., c^k\in\mathbb{R}^D$ are called control points. In the remainder of the paper, we assume that the linear part of the registration map \eqref{eq: model general} is restricted to scaling and rotation $B=SR$, where $S=\text{diag}(s_1,\ldots,s_D)$ with each $s_i>0$ is the scaling matrix, and $R\in\mathbb{R}^{D\times D}$ is the rotation matrix. Hence, we use the following registration map \footnote{In methods based on TPS (Thin-Plate Spline), as discussed in sub-sections B and D, we continue to apply the traditional TPS-regression technique. This approach typically recovers a general linear matrix $B$. As such, these methods do not guarantee a linear mapping in the form of $SR$ finally, unless the true deformation mapping between $X$ and $Y$ satisfies this assumption.}: 
\begin{align}
f(x):=\sum_{k=1}^K\alpha^k\phi(x,c^k)+R^TSx+\beta \label{eq: RBF model}.
\end{align} 

At their core, the proposed methods iteratively update the transformation parameters in two steps, estimating the correspondence and updating the transform parameters based on recent correspondence. For the first step, we will apply sliced optimal partial transport and optimal partial transport; for the second step, we will apply the RBF regression method and TPS regression methods. In particular, our methods will be introduced in the following four subsections. Additionally, as one condition in the problem setup, we assume $\zeta$ is the number of points of the ``clean part,'' which is given as our prior knowledge. We first introduce the main ideas of our methods in the following four subsections; then, in the last section, we summarize these methods as pseudo-codes before we proceed to our numerical experiments. 

Before we introduce our methods, we first formulate the point cloud registration with noise as the following problem:
\begin{align}
\min_{f}\min_{L} \sum_{n\in \text{Dom}(L)}\|f(x^n)-y^{L[n]}\|^2+\epsilon \text{reg}(f) \label{eq: registration object}
\end{align}
where $L$ is over all $[1:N]\hookrightarrow[1:M]$ partial bijections such that $|\text{Dom}(L)|=\zeta$. And there exists an underlying solution $L^*$, that satisfies the above conditions.  The $\text{reg}({f})$ term is a regularization term. When $f$ is RBF model \eqref{eq: RBF model} or TPS model \eqref{eq: TPS model}, we refer \eqref{eq: RBF obj} and \eqref{eq: TPS obj} for its particular formulation. 
\vspace{-1em}
\subsection{OPT-RBF}
Our paper's primary contribution is the application of sliced OPT to the non-rigid registration of point sets. To provide context and motivation for our approach, we begin by formulating the problem using the primal form of the partial transport, as given in Eq. \eqref{eq: OPT primal empirical}. 

Given two point sets $X=\{x^n\in\mathbb{R}^D\}_{n=1}^N$ and $Y=\{y^m\in\mathbb{R}^D\}_{m=1}^M$, we treat these sets as empirical measures, $\mu=\sum_{n=1}^N \delta_{\hat y^n}$ and $\nu=\sum_{m=1}^M \delta_{y^m}$. The OPT-RBF then can be formalized as the optimization problem in \eqref{eq: registration object}.
Note that this optimization problem is similar to the Wasserstein Procrustes problem \cite{grave2019unsupervised} but with two significant differences: 1) $f$ is non-rigid, and 2) we use OPT as opposed to OT. To solve this optimization problem, we propose an alternating iterative optimization scheme through the following two steps: 


\noindent\textbf{Step 1}. For fixed registration parameters, first calculate,
$$\hat{y}^n=f(x^n),\forall n\in[1:N],$$
where $f$ is defined by \eqref{eq: RBF model}. Regarding control point set $\{c^1, \ldots, c^K\}$, we can either set it as source point set $\{x^1,\ldots x^n\}$ or choose a different point set based on prior model knowledge.
Then we solve the primal OPT problem:
\begin{align}
  \gamma^*=\text{primal-OPT}(\{\hat {y}^n\}_{n=1}^N,\{y^n\}_{m=1}^M;\zeta)  \label{eq: solve opt}
\end{align}

where $\zeta$ is the number of points (the mass) of the clean data point cloud, and $\gamma^*$ is the optimal partial transportation plan. Let 
\begin{align}
  \mathcal{D}=\{n: \sum_{m=1}^M(\gamma_{n,m})> 0\}\label{eq: domain D OPT},
  \end{align}
  then for each $n\in\mathcal{D}$, we update $\hat{y}^n$ via: 
\begin{align}
\hat{y}^n\gets \left(\frac{1}{\sum_{m=1}^M(\gamma_{n,m})}\gamma Y\right)[n,:]. \label{eq: update Y_hat opt}  
\end{align}
which is the barycentric projection of the partial transport plan \cite{ambrosio2005gradient,bai2023linear}. Lastly, we obtain the following correspondence\footnote{See appendix \ref{sec:notations} for detailed explanation of the term ``correspondence''.}:
$$\{(x^n,\hat{y}^n)\}_{n\in\mathcal{D}}.$$

\noindent\textbf{Step 2.} Given a correspondence 
$\{(x^n,\hat{y}^n)\}_{n\in \mathcal{D}}$, we will update $\{\alpha^k\}_{k=1}^K, S,R,\beta$. 
For convenience, by reindexing, we suppose correspondence is $\{(x^n,\hat{y}^n)\}_{n=1}^{N_{sub}}$, where $N_{sub}\leq \min(N, M)$, i.e. $\mathcal{D}=[1:N_{sub}]$.

The parameters $S,R,\beta,\alpha$ are selected via to minimizing the following: 
\begin{align}  &\min_{S,R,\beta,\alpha}\sum_{n=1}^{N_{sub}}\left\|\left(\sum_{k=1}^K\alpha^k\phi(x^n,c^k)+R^TSx^n+\beta\right)-\hat y^n\right\|^2 \nonumber\\
&\quad\quad\quad+\epsilon\sum_{k=1}^K\|\alpha_k\|^2 \label{eq: RBF obj} 
\end{align}
where $\epsilon>0$, and the regularization term $\epsilon\sum_{i=1}^K\|\alpha_k\|^2$ is applied to restrict the model flexibility and improve the numerical computational stability.

Let 
\begin{align}
&X_{sub}=\left[\begin{array}{c}
    (x^1)^T\\ 
    \ldots\\
   (x^{N_{sub}})^T
\end{array}\right], 
\hat{Y}_{sub}=\left[\begin{array}{c}
    (\hat y^{1})^T\\ 
    \ldots\\
    (\hat y^{N_{sub}})^T
\end{array}\right], 
\alpha=\left[\begin{array}{c}
     (\alpha^1)^T  \\
     \ldots \\ 
     (\alpha^K)^T
\end{array}
\right],\nonumber\\ 
&\Phi=\left[\begin{array}{ccc}
\phi(x^1,c^1) &\ldots & \phi(x^1,c^K)  \\
\ldots     &  \\ 
\phi(x^{N},c^1) &\ldots & \phi(x^{N},c^K)
\end{array}\right],\Phi_{sub}=\Phi[\mathcal{D},:]. \nonumber 
\end{align}

Thus, \eqref{eq: RBF obj} can be written in the following matrix form:
\begin{align}
\min_{S,R,\beta,\alpha}\|\Phi_{sub}\alpha+X_{sub}SR+\beta^T1_{N_{sub}}-\hat{Y}_{sub}\|^2+\epsilon\cdot\text{tr}(\alpha^T\alpha)\label{eq: BRF obj2}
\end{align}

We will update the transformation parameters in two steps:

Step 2.1: Fix $\alpha$, and let $\hat{Y}'=\hat{Y}_{sub}-\Phi_{sub}\alpha$, that is, each entry, noted as $\hat{y}'^n=\hat{Y}'[n,:]^T$, is defined as 
$$\hat{y}'^n=\hat y^n-\sum_{k=1}^K\alpha^k\phi(x^n,c^k),\forall n\in[1:N_{sub}].$$
We aim to solve 
$$\min_{S,R,\beta}\sum_{n=1}^{N_{sub}}\|f(x^n)-\hat{y}^n\|^2=\|X_{sub}SR+\beta^T1_{N_{sub}}-Y'\|^2.$$
By \cite{du2007extension}, we can obtain the optimal $R, S,\beta$ as follows: 

Let $Y_c=\hat{Y}'-\left[\begin{array}{c}
     \bar{y}^T  \\
     \ldots\\
     \bar{y}^T
\end{array}\right] 
$ where $\bar{y}=\frac{1}{N_{sub}}\sum_{n=1}^{N_{sub}}(\hat y'^n)$ and $y_c^n=Y_c[n,:]^T$ denote $n^{th}$ entry of $Y_c$, for each $n$. Similarly, we define $X_c$ and $x_c^n$. 

Given a scaling matrix $S$ (initially, $S$ is set to be identity matrix $I_D$), let 
\begin{align}
H_c=(X_cS)^TY_c=U_{H}\Sigma_H V_H^T\label{eq: H_c}.
\end{align}
The optimal $R$ (for fixed $S$) is given by 
\begin{align}
    R=U_H  \left[\begin{array}{cc}
    I_{D-1} & 0 \\ 
    0 & \text{det}(V_H U_H^T)
    \end{array}\right]V_H^T
    \label{eq: optimal R} 
\end{align}

Next, given $R$, the optimal $S=\text{diag}(s_1,\ldots,s_d)$ is computed by 
\begin{align}
  s_d=\frac{\sum_{i=1}^{N_{sub}} y^{iT}_cR^TE_dx^i_c}{\sum_{i=1}^{N_{sub}}x^{iT}_cE_dx^{i}_c},\forall d  \label{eq: optimal S}
\end{align}
  where $E_d=[0,\ldots, e_d, \ldots,0]$ and $e_1,...e_D$ are the Canonical vectors in $\mathbb{R}^D$.
We iterate through steps \eqref{eq: optimal R} and \eqref{eq: optimal S} until $R$ and $S$ converge. Note, in the case of uniform scaling, where $S = sI_D$ for some $s > 0$, it is sufficient to compute \eqref{eq: optimal R} just once to find the optimal values for $R$. Additionally, the scaling factor 
$S=sI_D$ can be calculated in two ways. One approach is to use the average value of 
$\{s_1,\ldots s_d\}$, as specified by Equation \eqref{eq: optimal S}. Alternatively, 
$s$ can be determined using the formula:
$s=\sqrt{\frac{\text{cov}(Y_c)}{\text{cov}(X_c)}}.$

Finally, the optimal translation $\beta$ is computed by: 
\begin{align}
\beta&=\frac{1}{N_{sub}}\sum_{n=1}^{N_{sub}}[(\hat{y}'^n-R^TSx^n)]\nonumber\\
&=\frac{1}{N_{sub}}[1_N^T\hat{Y}'-1_N^TX_{sub}SR]\label{eq: optimal beta}
\end{align}

Step 2.2: Given $S,R,\beta$, we find the optimal $\alpha$. 
Let $\hat{Y}''=Y-X_{sub}SR-\beta^T1_N$, we have 
\begin{align}
&\arg\min_{\alpha}\|\Phi_{sub}\alpha+X_{sub}SR+\beta^T1_N-\hat{Y}_{sub}\|^2+\epsilon\text{tr}(\alpha^T\alpha)\nonumber\\
&=\arg\min_{\alpha}\|\Phi_{sub}\alpha-\hat{Y}''\|^2+\epsilon \text{tr}(\alpha^T\alpha) \nonumber \\
&=\arg\min_{\alpha}\sum_{d=1}^D\|\Phi_{sub}\alpha[:,d]-\hat{Y}''[:,d]\|^2+\epsilon \alpha[:,d]^T\alpha[:,d] \nonumber 
\end{align}
It suffices to solve the following 1D regression problem:
\begin{align}
\min_{\alpha[:,d]} \|\Phi_{sub}\alpha[:,d]-\hat{Y}''[:,d]\|^2+\epsilon \alpha[:,d]^T\alpha[:,d]   \label{eq: 1D RBF obj}
\end{align}
 for each $d\in[1:D]$, whose solution is given by 
\begin{align}
\alpha[:,d]=(\Phi^T_{sub}\Phi_{sub}+\epsilon I_K)^{-1}\Phi^T_{sub}\hat{Y}''[:d] \nonumber 
\end{align}
Therefore, optimal $\alpha$ is given by 
\begin{align}
\alpha=(\Phi^T_{sub}\Phi_{sub}+\epsilon I_K)^{-1}\Phi^T_{sub}\hat{Y}''.\label{eq: optimal alpha} 
\end{align}
We repeat these two steps until convergence. 

\vspace{-1em}
\subsection{OPT-TPS}
The second method we introduced is the combination of OPT and TPS, it can be regarded as a variant of TPS-RPM \footnote{For further details, please see Appendix \ref{sec:Sinkhorn}.}. In particular, we use the TPS model:
\begin{equation}
f(x)=\sum_{n=1}^N\alpha^n\phi_T(x,x^n)+B^Tx+\beta \label{eq: TPS model}
\end{equation}
Same to the \textbf{Step 1} in OPT-RBF, we use primal-OPT to estimate the correspondence and obtain $\{(x^n,\hat{y}^n)\}_{n=1}^N$. It is important to note that in this method, we utilize the complete set of $x$ and $\hat{y}$ points. 

Next, we discuss how to update the parameters of the above model based on a given correspondence. We first review the TPS regression technique.

\textbf{Introduction of TPS regression in $\mathbb{R}^D$.} 

As we introduced in Section \ref{sec:related_work}, the thin-plate spline is derived from the following minimization problem,  
\begin{equation}
\inf_{\tilde{f}:\mathbb{R}^D\to\mathbb{R}}E[\tilde{f}]:=\inf_{\tilde{f}:\mathbb{R}^D\to\mathbb{R}}\sum_{n=1}^N\|\tilde{f}(x^n)-\tilde{y}^n\|^2+\epsilon \int_{\mathbb{R}^D}\|\nabla^2\tilde{f}\|^2dx  \label{eq: 1D TPS obj}.
\end{equation}
where $\| \nabla^2\tilde{f}\|^2=\sum_{i,j=1}^D\left(\frac{\partial^2\tilde{f}}{\partial x_i\partial x_j}\right)^2$, the regularization term is called \textbf{bending energy};
$\{x^n\}_{n=1}^{N}\subset\mathbb{R}^D$ and $\{\tilde{y}^n\}_{n=1}^{N}\in\mathbb{R}$ are given fixed data points. 

It has been proved that the solution $\tilde{f}$ for above \eqref{eq: 1D TPS obj} has the following closed form: 
\begin{align}
\tilde{f}(x)&
=\sum_{n=1}^{N}\alpha^n\phi_{T}(x,x^n)
+b^Tx+\tilde\beta\label{eq: TPS function}
\end{align}
where $\tilde\beta,\alpha^n \in\mathbb{R},b\in\mathbb{R}^n$ and 
\begin{align}
 \phi_T(x,x^n):=\begin{cases}
 \frac{1}{2}\|x-x^n\|^2 \ln (\|x-x^n\|^2) & \text{if }D=2 \\ 
  \ln (\|x-x_0\|) & \text{if }D=4 \\
  \|x-x_0\|^{4-D} & \text{otherwise}
 \end{cases} \label{eq: TPS kernel}. 
\end{align}
\begin{remark}
In practice, when $D\ge 4$, $\phi_T$ has singularity at $x=x^n$ and thus $\phi_T(x,x^n)$ is set to be $\|x-x^n\|^2$ or $\|x-x^n\|^2\ln(\|x-x^n\|^2)$. Therefore the constructed $f(x)$ is generally not the minimizer for \eqref{eq: 1D TPS obj}. 
\end{remark}
Let
\begin{align}
\bar{X}=\left[\begin{array}{c}
     1,x^1 \\
     \cdots\\ 
     1,x^{N}
\end{array}\right]=[1_N,X],
\tilde{Y}=\begin{bmatrix}
    \tilde{y}_1\\ 
    \ldots\\ 
    \tilde{y}^{N}
\end{bmatrix},
\bar{b}=\left[\begin{array}{c}
\tilde\beta \\
b
\end{array}\right].\nonumber
\end{align}
The optimal parameter $\alpha, \bar{b}$ can be constructed by solving the following linear system 
\begin{align}
\begin{cases}
&\tilde{Y}=(\Phi+\epsilon I_{N})\alpha+\bar{X}\bar{b} \\     
&\bar{X}^T\alpha=0_N    
\end{cases} \label{eq: TPS regression}
\end{align}
where the second equation follows from the functional analysis in \cite{meinguet1979multivariate}, which force the $\alpha$ be in the null space of $\bar{X}$. Problem
\eqref{eq: TPS regression} can be solved by the following linear system 
\begin{equation}
\begin{bmatrix}
    \Phi+\epsilon I_N & \bar{X} \\ 
    \bar{X}^T & 0 
\end{bmatrix} \begin{bmatrix}
    \alpha \\ 
    \bar{b}
\end{bmatrix}=\begin{bmatrix}
    \tilde{Y}\\ 
    0
\end{bmatrix}\label{eq: TPS regression2}
\end{equation}
Thus 
\begin{align}
    \begin{bmatrix}
        \alpha\\ 
        \bar{b} 
\end{bmatrix}=\begin{bmatrix}
    \Phi+\epsilon I_N & \bar{X} \\ 
    \bar{X}^T & 0 
\end{bmatrix}^{-1}\begin{bmatrix}
    \tilde{Y}\\ 
    0
\end{bmatrix}. \label{eq: TPS solution1}
\end{align}
Alternatively, the matrix decomposition technique can be applied, and thus we have optimal parameter $(\alpha,\bar{b})$ has the following closed form: 
\begin{align}
\begin{cases}
&\alpha=(\Phi+\epsilon I_N)^{-1}(\tilde{Y}-\bar{X}\bar{b}),\\ &\bar{b}=(\bar{X}^T(\Phi+\epsilon I_N)^{-1}\bar X)^{-1}\bar{X}^T(\Phi+\epsilon I_N)^{-1}\tilde{Y}.     
\end{cases}
\label{eq: TPS solution2}
\end{align}
However, both \eqref{eq: TPS solution1}\eqref{eq: TPS solution2} are not computational practicable and \cite{wahba1990spline} proposed an improved computational method: 

We first write the QR decomposition of $\bar{X}$: 
$$
\bar{X}=[Q_1, Q_2] \begin{bmatrix}
    \mathcal{R}\\ 
    0
\end{bmatrix},
$$
where $Q_1,Q_2$ are $(D+1)\times N, (N-D-1)\times N$ orthogonal matrices; and $R\in \mathbb{R}^{(D+1)\times (D+1)}$ is upper triangular. 

\begin{align}
\begin{cases}
&\alpha=Q_2(Q_2^T(\Phi+\epsilon I_N)Q_2)^{-1}Q_2^T\tilde{Y} \\ &\bar{b}=\mathcal{R}^{-1}Q_1^T(\tilde{Y}-(\Phi+\epsilon I_N)\alpha)
\end{cases}.
\label{eq: TPS solution3}
\end{align}

\textbf{Step 2: TPS method regression for point cloud registration}.

Based on the above TPS interpolation technique, given a correspondence $(\{(x^n,\hat{y}^n)\}_{n=1}^N)$, the $D-$dimensional TPS regression can be simplified to be the following:
\begin{align}
&\inf_{\alpha,B,\beta}\sum_{n=1}^N\|f(x^n)-\hat{y}^n\|^2+\epsilon \sum_{d=1}^D\int_{\mathbb{R}^D}\|\nabla^2f[d]\|^2dx \label{eq: TPS obj}.  \end{align}

The optimal parameters $\alpha, B, \beta$ can be computed by solving the 1D TPS interpolation problem 
$$\inf_{\alpha[:,d],B[:,d],\beta[d]}\sum_{n=1}^N\|f(x^n)[d]-\hat{y}^n[d]\|^2+\epsilon \int_{\mathbb{R}^D}\|\nabla^2f[d]\|^2dx
$$
for each $d\in[1:D]$.
By \eqref{eq: TPS solution1} \eqref{eq: TPS solution2}, let 
$$\bar{B}=\begin{bmatrix}
\beta^T\\B
\end{bmatrix},\hat{Y}=\begin{bmatrix}\hat y^{1T}\\
\ldots\\ 
\hat y^{NT}
    \end{bmatrix}
,$$ 
we have 
\begin{align}
\begin{cases}
&\alpha=Q_2(Q_2^T(\Phi+\epsilon I_N)Q_2)^{-1}Q_2^T\hat{Y}\\ 
&\bar{B}=\mathcal{R}^{-1}Q_1^T(\hat{Y}-(\Phi+\epsilon I_N)\alpha).    
\end{cases}
 \label{eq: TPS solution4}
\end{align}

\vspace{-1em}
\subsection{SOPT-RBF methods}
In this method, we use the RBF modle \eqref{eq: RBF model}. 
As we discussed in \ref{sec:background}, computing OPT can be time consuming. In improve the computational efficiency. We use Sliced-OPT to replace the primal-OPT in step 1 in method OPT-RBF. 

\textbf{Step 1}. We first select $\{\theta_1,\theta_2,\ldots \theta_{t}\}\subset\mathbb{S}^{D-1}$. 
For $t'=1,2,\ldots, t$, we project $Y$ and $\hat{Y}$ into 1D space spanned by $\theta_{t'}$ and obtain 
$$\hat Y_{\theta_{t'}}:=\{\theta_{t'}^T\hat{y}^n\}_{n=1}^N,Y_{\theta_{t'}}=\{\theta_{t'}^T y^{m}\}_{m=1}^M,$$ and then we solve the 1D OPT problem 
\begin{align}
  \text{OPT}_{\lambda}(\hat{Y}_{\theta_{t'}},Y_{\theta_{t'}}). \label{eq: solve 1D OPT}  
\end{align}
Let $L_{t'}$ denote the OPT transportation map,\footnote{In OPT-RBF method, we use transportation matrix $\gamma$ to represent the transportation plan. While we use Monge mapping $L$ to denote the transportation plan in this section. These two descriptions are equivalent. For convenience, we use these two descriptions interchangeably. }. then we update $\hat{y}^n$ with: 
\begin{align}
  \hat{y}^n\gets \hat{y}^n+ \theta_{t'}(\theta_{t'}^Ty^{L_{t'}(n)}-\theta_{t'}^T\hat{y}^n). \label{eq: update y_hat sopt} 
\end{align}
We repeat the above process for $\theta_1,\theta_2,\ldots \theta_{t}$. 

Let $\mathcal{D}_{t'}:=\text{Dom}(L_{t'})$, and $\mathcal{D}=\bigcup_{t'} \mathcal{D}_{t'}$, that is, $\mathcal{D}$ is union of index of $\hat{y}^n$ which has been moved in above process. We have 
$\{({x}^n,\hat{y}^n)\}_{n\in\mathcal{D}}$ defining a correspondence. 
For step 2, it is same to the \textbf{step 2} in OPT-RBF. 


\subsection{SOPT-TPS}
This section is an SOPT version of the OPT-TPS method. In particular, we use TPS model \eqref{eq: TPS model}. 

In particular, we apply the \textbf{Step 1} in SOPT-RBF to estiamte the correspondence and use \textbf{Step 2} in OPT-TPS to update the parameter. 
\vspace{-1em}
\subsection{Summary and pseudo-code}
\SetKwComment{Comment}{/* }{ */}
\SetKwInput{KwInput}{Input}
\SetKwInput{KwOutput}{Output}
\vspace{-0em}
\begin{algorithm}
\caption{OPT-RBF}\label{alg:opt-rbf}
\KwInput{$X,Y,\zeta,T,\phi,\text{param},\epsilon,C\gets X$}
\KwOutput{$R,S,\beta,\Phi,\alpha$}
\text{Initilize} $R,S,\beta,\alpha$\\
\text{Initilize} $\Phi\gets \phi(X,C^T,\text{param})$\\
\For{$T'=1,2,\ldots T$}{
\Comment{Step 1}
$\hat{Y}\gets\Phi\alpha+XSR+\beta^T1_N$\\ 
compute optimal plan $\gamma$ for \text{Primal-OPT}$(\hat{Y},Y;\zeta)$. \\ 
$\mathcal{D}\gets\{n:\sum_{m=1}^M\gamma_{n,m}>0\}$\\ 
update $\hat{y}^n,\forall n\in \mathcal{D}$ via \eqref{eq: update Y_hat opt} \\ 
$(X_{sub},\hat{Y}_{sub},\Phi_{sub})\gets (X[\mathcal{D},:],\hat{Y}[\mathcal{D},:],\Phi[\mathcal{D},:])$\\
\Comment{Step 2}
$\hat{Y}'\gets \hat{Y}_{sub}-\Phi_{sub}\alpha$\\
By $(X_{sub},\hat{Y}')$, update $R,S,\beta$ via \eqref{eq: optimal R},\eqref{eq: optimal S},\eqref{eq: optimal beta}. \\

\If{condition for non-rigid is True}{
$\hat{Y}''\gets \hat{Y}_{sub}-(X_{sub}SR+\beta^T 1_{N_{sub}})$\\ 
By $(\Phi_{sub},\hat{Y}'')$, update $\alpha$ via \eqref{eq: optimal alpha}
}
}
\end{algorithm}
The pseudo-code for OPT-RBF is presented in Algorithm \ref{alg:opt-rbf}. For the inputs, $X,Y$ are source and target point clouds, $\zeta$ is the number of clean data points of $X$, which is the prior knowledge and has been elaborated upon in Section \ref{sec:method}. The variable 'param' represents the kernel-specific parameters; for example, $\sigma^2$ in the case of the Gaussian kernel and dimension $D$ for the TPS (Thin-Plate Spline) kernel. $C=[c^1,\ldots c^K]^T\in\mathbb{R}^{K\times D}$ is the set of control points, where $K\in\mathbb{N}$. By default, $C$ is set to $X$, but it can be configured to other sets of points represented as matrices.

Initialization occurs at line 1, where we conventionally set $R=I_D$, $S=I_D$, $\beta=0_{D}$, and $\alpha=0_{N \times D}$. Line 11 sets forth a condition to initiate the nonrigid registration process. Typically, this condition is established as enough rigid iterations, coupled with the convergence of the linear parameters, $R,S,\beta$, or $\bar{B}$. 

For the pseudo-code of the other three algorithms, please refer to Appendix \ref{sec:code}.

\begin{remark}
In a balanced sliced OT setting, the steps specified by \eqref{eq: update y_hat sopt} is formally known as \textbf{sliced Wasserstein gradient flow} \cite{figalli2021invitation, liutkus2019sliced}. As shown in \cite[Theorem 1.1, Theorem 4.7]{li2023measure}, the transported point set $\hat{Y}$ converges to $Y$ as the number of iterations tends to infinity. However, convergence properties in the partially sliced OT setting are not yet fully understood. Empirical observations indicate rapid convergence when the rotation angles between $X$ and $Y$ are smaller than $\pi/2$. A comprehensive theoretical analysis of this phenomenon is a subject for future research.
\end{remark}

\begin{remark}\label{rm: sot}
As we discussed in sub-section A, under the balanced OT setting, if we exclude the non-rigid part ``$\Phi\alpha$'', the problem formulation given by \eqref{eq: registration object} is known as the \textbf{Wasserstein Procrustes problem}. As per \cite{grave2019unsupervised}, the gradient technique with full sample points faces challenges related to statistical convergence and computational expense. The computational expense of solving OT(or OPT) has been discussed in section \ref{sec:background}. Regarding convergence, the Wasserstein Procrustes is a non-convex problem.\footnote{To navigate this, one should reinterpret $L$, equivalent to a permutation matrix, as any bi-stochastic matrix. This modification makes the search space for the variable $L$ convex. Thus, the discussion of convexity becomes suitable.} In \cite{grave2019unsupervised}, the benefits of stochastic gradient descent are explored. sliced OT or (OPT) can intuitively produce similar results. By reducing $X, Y$ to a one-dimensional domain, partial information is lost. As a result, the correspondence derived is generally different from the correspondence from the original D-dimensional OT (or OPT) approach, and randomness has been introduced. Integrating with the partial OT setting, only the points in $\mathcal{D}$ are transported, which is similar to the sub-sampling step in gradient descent. This intuition aligns with our experiment observation, and we aim to rigorously evaluate the advantages of the sliced unbalanced OT method concerning the Wasserstein Procrustes problem in future works.
\end{remark}

\vspace{-0.5em}
\section{Experiments}\label{sec:experiments}
\begin{table*}

\begin{center}
\textbf{Wall clock time (in second) and registration error for all methods}\\
\vspace{1em}
\begin{tabularx}
{1.0\textwidth}{| c | c | C | C | C | C | C | C |C| c|}
\hline
 &  & \multicolumn{2}{c|}{female}  & \multicolumn{2}{c|}{male} & \multicolumn{2}{c|}{neutral}&\multicolumn{2}{c|}{registration error }\\
\cline{3-10}
 & & $5\%$ & $10\%$ & $5\%$ & $10\%$ & $5\%$ & $10\%$ & $5\%$ & $10\%$ \\
\hline
 \multirow{7}{*}{\rotatebox[origin=c]{90}{baselines}}
& CPD \cite{myronenko2010point}
& $5.44$& $5.86$ 
& $5.37$ & $6.82$ & $5.35 $  & $6.74 $ &0.063(0.022) &0.362(0.424)\\ 
 & OT-RBF 
 &$10.76$ &$11.70$
 &$10.13$ & $11.90$
 &$9.50$ & $10.33$ &0.339(0.112) &0.550(0.040)\\
 & OT-TPS 
 &$13.81$ &$14.01$
 &$13.26$ & $14.33$
 &$12.35$ & $14.21$&0.076(0.012)& 0.151(0.044)\\
& SOT-RBF 
&$3.47$ &$3.50$ 
&$3.43$ & $3.49$ &$3.36$ & $3.49$ & 0.121(0.062)&0.219(0.185)
\\
& SOT-TPS 
& $6.38$ &$7.52$ 
&$6.43$ & $7.53$ 
&$6.41$ & $7.58$ &0.046(0.007)& 0.065(0.004)\\

& TPS-RPM\cite{chui2003new} 
 &$15.1$&$16.6$ 
 &$15.2$ & $17.6$
 &$14.8$ &$16.4$  & 0.183(0.011)&0.218(0.018)\\ 
 
 & TPS-RPM(new) 
 &$14.5$&$16.7$ 
 &$15.5$ &$17.7$ 
 &$15.3$ &$17.8$ & 0.039(0.010) &0.041(0.010)\\ 
\hline 
\multirow{4}{*}{\rotatebox[origin=c]{90}{\makebox[1cm][c]{ours}}}
 & OPT-RBF& 
 $11.43$ & $79.1$ 
 &$11.53$ & $72.9$ 
 &$11.11$ &$73.8$ &$0.037(0.008)$  & $\textbf{0.040(0.011)}$\\
 
 & OPT-TPS& 
 $14.91$ & $74.6$ &
 $15.45$ & $72.4$ & 
 $15.40$ & $64.4$ & $\textbf{\textbf{0.033(0.010)}}$ & $\textbf{0.033(0.010)}$\\
 
& SOPT-RBF
&$8.37$ &$9.90$ 
&$8.42$ & $9.55$ 
&$9.24$ & $9.93$& $\textbf{0.036(0.013)}$ &$0.042(0.011)$\\ 
& SOPT-TPS
 &$11.71$ & $12.21$ 
 &$11.59$ & $12.72$ 
 &$11.55$ &$12.31$ & $\textbf{\textbf{0.034(0.009)}}$ & $\textbf{0.040(0.009)}$\\

\hline 
\end{tabularx}
\end{center}
\caption{In this table, the first six columns display the wall clock time for each method, measured in seconds per iteration. All methods in our experiment converge within a range of 45 to 65 iterations. The final two columns show the registration error for each method. In these cells, the first value represents the mean registration error, where the error is defined in \ref{eq: error}, while the value in parentheses indicates the standard deviation.}
\label{tb:time}
\end{table*}

In this experiment, the methods being compared are our methods (see section \ref{sec:method}): \textbf{SOPT-RBF}, \textbf{SOPT-TPS}, \textbf{OPT-RBF}, \textbf{OPT-TPS}; and baseline methods: \textbf{CPD} \cite{myronenko2010point},  \textbf{TPS-RPM} \cite{chui2003new,yang2011thin},and an improved version of TPS-RPM (See Appendix \ref{sec:Sinkhorn} for more details), denoted as \textbf{TPS-RPM(new)}. In addition,  we extend the OT Procrustes method\cite{grave2019unsupervised} into the non-rigid setting by our RBF and TPS models, and denoted as \textbf{OT-RBF}, \textbf{OT-TPS}. Furthermore, we provide the sliced-OT version, denoted as \textbf{SOT-RBF} and \textbf{SOT-TPS}. These five methods are baselines for the experiments.

The dataset being used is \textit{STAR dataset} (\url{https://github.com/ahmedosman/STAR}). This dataset contains a list of point clouds with labels ``female'', ``male'', and ``neutral''. For each label, we randomly select two datasets, denoted as $X$ and $Y$ as the source and target datasets, respectively.

The support of all point clouds is $[-1,1]^3$. For each pair of source and target point clouds, the rotation angle in each dimension is set to be in the range 
$[-\frac{\pi}{3},\frac{\pi}{3}]$. Additionally, the translation $\beta$ is selected such that$|\beta[d]|\leq 1$ for $d\in[1:3]$. Scaling is set to be the identity matrix and remains unchanged for all methods.

Both the source and target datasets have uniform noise added to them, which is distributed in the support $[-1, 1]^3$. This experiment compares the accuracy and running time performance of the mentioned methods under these conditions.

\textbf{Accuracy}. 
Suppose $X,Y$ are the source and target data respectively and $X_0\subset X, Y_0\subset Y$ be the clean part of $X, Y$. Say $L^*,f^*$ is the ground true correspondence and deformation function, i.e. 
$$y^{L^*(n)}=f^*(x^{n}),\forall x^n\in X_0.$$
Suppose $\zeta$ is the size of $Y_0$. Let $\sigma(Y_0)$ be the standard deviation of $Y_0$, i.e. $\sigma(Y_0)>0$ such that
$\frac{1}{D}\frac{1}{N}\left\|\frac{Y_0-\text{mean}(Y_0)}{\sigma(Y_0)}\right\|^2=1.$
For each method, let function $f$ be the returned model, the approximation error is defined as 
\begin{align}
  \text{error}=\left(\frac{1}{\zeta}\sum_{x^n\in X_0}\left\|\frac{y^{L^*(n)}-f(x^{n})}{\sigma(Y_0)}\right\|^2\right)^{1/2} \label{eq: error}  
\end{align}

\begin{figure}
    \centering
\includegraphics[width=0.75\columnwidth]{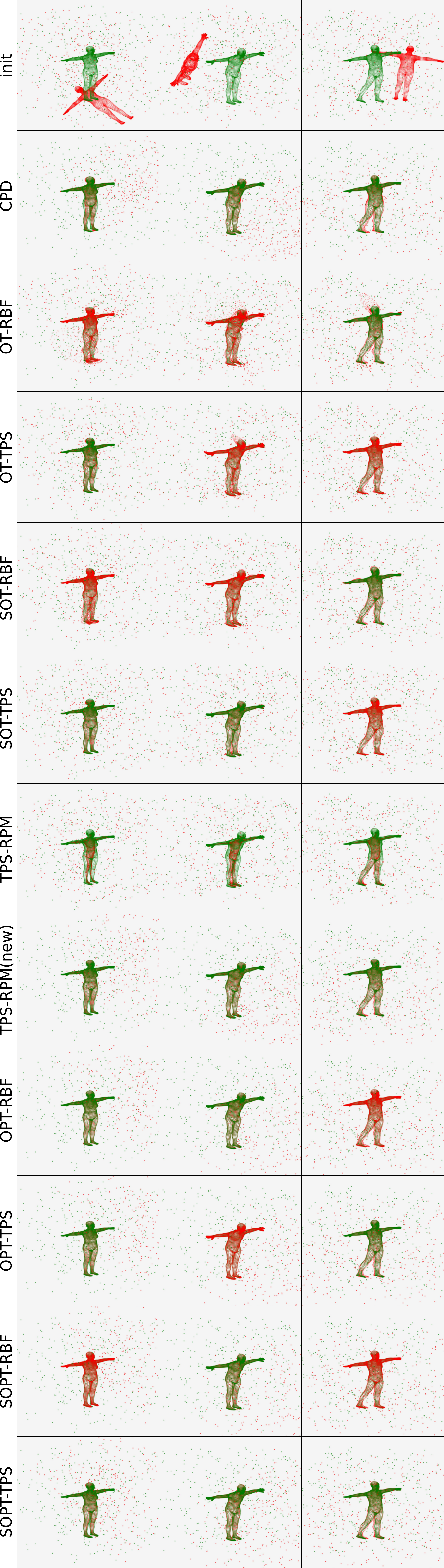}
    \caption{This figure visually represents the final result obtained from all methods, all under a noise level of 10\%. }
    \label{fig:3D result}
\end{figure}

For each method, parameters are carefully selected for optimal performance. Please refer to the following section on \textbf{Performance Analysis} for a detailed introduction. Table \ref{tb:time} and Figure \ref{fig:3D result} reveal that our methods, along with TPS-RPM(new), consistently yield superior accuracy, thanks in part to the utilization of prior knowledge $\zeta$. Additionally, sliced-OT methods, SOT-RBF and SOT-TPS, also contribute to improved accuracy. This aligns with our intuitive understanding presented in Remark \ref{rm: sot}, which suggests that the sliced-OT approach introduces randomness in the corresponding estimation step. This proves advantageous in the registration process, particularly when the correspondence established by balanced OT is inaccurate due to random noise.



\textbf{Performance analysis}. We evaluated the wall-clock time for our methods alongside the four baseline techniques, the results of which can be found in Table \eqref{tb:time}.
This table displays the wall-clock time per iteration and the total number of iterations needed for convergence for each method and dataset.

All the methods are configured to run for a maximum of 100 iterations, and rigid registrations are performed exclusively within the initial 20. Convergence is generally observed between 45 and 65 iterations. 

In the CPD method, the Gaussian parameter term $\sigma^2$ is fixed at 16.0. The scaling term is set to 1.0, and the weight term used for the non-rigid M-step in linear regression is fixed at 4.0. 

In TPS-RPM and TPS-RPM (new), the Sinkhorn algorithm \cite{cuturi2013sinkhorn} is accelerated by using Numba (\url{https://numba.pydata.org/}). 
We've set the weight of entropy regularization to $\sigma(Y_0)*0.01$ and capped the maximum number of iterations at 500 to prioritize the performance.  As for OPT-RBF and OPT-TPS, we utilize the OT solver from PythonOT \cite{flamary2021pot}, a C++ library, and impose a maximum of 1e7 iterations.  Regarding SOPT-RBF and SOPT-TPS, we set the number of projections to be 100 and the 1D-OPT solver  \cite{bai2022sliced} is accelerated via Numba \url{https://numba.pydata.org/}. For all these methods, steps involving the computation of optimal non-rigid parameters, particularly those requiring the inverse of a large-size matrix, are executed on a GPU. 
 The data type of each dataset is the 64-bit float number, and all the experiments are conducted on a machine with an AMD EPYC 7713 64-core Processor and four NVIDIA RTX A6000 GPUs. The CPU operates at a maximum of 3720.7 MHz and supports multi-threading with 128 threads.

\vspace{-1.5em}
\section{Summary}\label{sec:summary}
In this paper, we proposed new non-rigid registration methods for partial matching scenarios by incorporating optimal partial transport and classical non-rigid registration models, RBF and TPS. In addition, to improve the computation efficiency, we also propose sliced-OPT-based methods. We demonstrate our methods in 3D and 2D datasets. In our experiment, when data is corrupted by some proportion of noise data, the partial OT-based method induces better accuracy. For future research, we plan to investigate the convergence behavior of methods based on OPT and sliced OPT. Moreover, we aim to explore the potential benefits of using a sliced-OT-based approach in addressing Wasserstein Procrustes problems, as suggested by our findings in both 3D and 2D experiments. Additionally, we intend to examine the potential applications of generalized sliced unbalanced OT to this problem.

\section*{Acknowledgement}
SK acknowledges partial support from the Defense Advanced
Research Projects Agency (DARPA) under Contract No. HR00112190135 and HR00112090023, and the Wellcome LEAP Foundation. SBD thanks the American Mathematical Society and Princeton University for financial support.
He also thanks Charles Fefferman and Keaton Hamm for many discussions related to `slow twists' and `slides.'

\bibliographystyle{IEEE-TIP/ieee_fullname}
\bibliography{IEEE-TIP/references}

\begin{thebibliography}{100}\itemsep=-1pt

\bibitem{abdelmunim2011elastic}
Hossam Abdelmunim and Aly~A Farag.
\newblock Elastic shape registration using an incremental free form deformation
  approach with the icp algorithm.
\newblock In {\em 2011 Canadian Conference on Computer and Robot Vision}, pages
  212--218. IEEE, 2011.

\bibitem{aguilar20173d}
Fernando~J AGUILAR, Ismael FERN{\'A}NDEZ, Juan~A CASANOVA, Francisco~J RAMOS,
  Manuel~A AGUILAR, Jos{\'e}~L BLANCO, and Jos{\'e}~C MORENO.
\newblock 3d coastal monitoring from very dense uav-based photogrammetric point
  clouds.
\newblock In {\em Advances on Mechanics, Design Engineering and Manufacturing:
  Proceedings of the International Joint Conference on Mechanics, Design
  Engineering \& Advanced Manufacturing (JCM 2016), 14-16 September, 2016,
  Catania, Italy}, pages 879--887. Springer, 2017.

\bibitem{alhashim2015deformation}
Ibraheem Alhashim, Kai Xu, Yixin Zhuang, Junjie Cao, Patricio Simari, and Hao
  Zhang.
\newblock Deformation-driven topology-varying 3d shape correspondence.
\newblock {\em ACM Transactions on Graphics (TOG)}, 34(6):1--13, 2015.

\bibitem{alismail2014continuous}
Hatem Alismail, L~Douglas Baker, and Brett Browning.
\newblock Continuous trajectory estimation for 3d slam from actuated lidar.
\newblock In {\em 2014 IEEE International Conference on Robotics and Automation
  (ICRA)}, pages 6096--6101. IEEE, 2014.

\bibitem{ambrosio2005gradient}
Luigi Ambrosio, Nicola Gigli, and Giuseppe Savar{\'e}.
\newblock {\em Gradient flows: in metric spaces and in the space of probability
  measures}.
\newblock Springer Science \& Business Media, 2005.

\bibitem{aoki2019pointnetlk}
Yasuhiro Aoki, Hunter Goforth, Rangaprasad~Arun Srivatsan, and Simon Lucey.
\newblock Pointnetlk: Robust \& efficient point cloud registration using
  pointnet.
\newblock In {\em Proceedings of the IEEE/CVF conference on computer vision and
  pattern recognition}, pages 7163--7172, 2019.

\bibitem{bai2023linear}
Yikun Bai, Ivan~Vladimir Medri, Rocio~Diaz Martin, Rana Shahroz, and Soheil
  Kolouri.
\newblock Linear optimal partial transport embedding.
\newblock In {\em International Conference on Machine Learning}, pages
  1492--1520. PMLR, 2023.

\bibitem{bai2022sliced}
Yikun Bai, Bernard Schmitzer, Mathew Thorpe, and Soheil Kolouri.
\newblock Sliced optimal partial transport.
\newblock {\em arXiv preprint arXiv:2212.08049}, 2022.

\bibitem{benamou2015iterative}
Jean-David Benamou, Guillaume Carlier, Marco Cuturi, Luca Nenna, and Gabriel
  Peyr{\'e}.
\newblock Iterative bregman projections for regularized transportation
  problems.
\newblock {\em SIAM Journal on Scientific Computing}, 37(2):A1111--A1138, 2015.

\bibitem{besl1992method}
Paul~J Besl and Neil~D McKay.
\newblock Method for registration of 3-d shapes.
\newblock In {\em Sensor fusion IV: control paradigms and data structures},
  volume 1611, pages 586--606. Spie, 1992.

\bibitem{Bonneel2019sliced}
Nicolas Bonneel and David Coeurjolly.
\newblock {SPOT}: sliced partial optimal transport.
\newblock {\em ACM Transactions on Graphics}, 38(4):1--13, 2019.

\bibitem{bonneel2015sliced}
Nicolas Bonneel, Julien Rabin, Gabriel Peyr{\'e}, and Hanspeter Pfister.
\newblock Sliced and {Radon} {Wasserstein} barycenters of measures.
\newblock {\em Journal of Mathematical Imaging and Vision}, 51(1):22--45, 2015.

\bibitem{bonneel2011displacement}
Nicolas Bonneel, Michiel Van De~Panne, Sylvain Paris, and Wolfgang Heidrich.
\newblock Displacement interpolation using lagrangian mass transport.
\newblock In {\em Proceedings of the 2011 SIGGRAPH Asia conference}, pages
  1--12, 2011.

\bibitem{bookstein1989principal}
Fred~L. Bookstein.
\newblock Principal warps: Thin-plate splines and the decomposition of
  deformations.
\newblock {\em IEEE Transactions on pattern analysis and machine intelligence},
  11(6):567--585, 1989.

\bibitem{brown2007global}
Benedict~J Brown and Szymon Rusinkiewicz.
\newblock Global non-rigid alignment of 3-d scans.
\newblock In {\em ACM SIGGRAPH 2007 papers}, pages 21--es. Association for
  Computing Machinery, 2007.

\bibitem{caffarelli2010free}
Luis~A Caffarelli and Robert~J McCann.
\newblock Free boundaries in optimal transport and monge-ampere obstacle
  problems.
\newblock {\em Annals of mathematics}, pages 673--730, 2010.

\bibitem{chapel2020partial}
Laetitia Chapel, Mokhtar~Z Alaya, and Gilles Gasso.
\newblock Partial optimal tranport with applications on positive-unlabeled
  learning.
\newblock {\em Advances in Neural Information Processing Systems},
  33:2903--2913, 2020.

\bibitem{chen2017matricial}
Yongxin Chen, Tryphon~T Georgiou, Lipeng Ning, and Allen Tannenbaum.
\newblock Matricial wasserstein-1 distance.
\newblock {\em IEEE control systems letters}, 1(1):14--19, 2017.

\bibitem{chen1992object}
Yang Chen and G{\'e}rard Medioni.
\newblock Object modelling by registration of multiple range images.
\newblock {\em Image and vision computing}, 10(3):145--155, 1992.

\bibitem{chizat2018scaling}
Lenaic Chizat, Gabriel Peyr{\'e}, Bernhard Schmitzer, and Fran{\c{c}}ois-Xavier
  Vialard.
\newblock Scaling algorithms for unbalanced optimal transport problems.
\newblock {\em Mathematics of Computation}, 87(314):2563--2609, 2018.

\bibitem{chizat2018unbalanced}
Lenaic Chizat, Gabriel Peyr{\'e}, Bernhard Schmitzer, and Fran{\c{c}}ois-Xavier
  Vialard.
\newblock Unbalanced optimal transport: Dynamic and {Kantorovich} formulations.
\newblock {\em Journal of Functional Analysis}, 274(11):3090--3123, 2018.

\bibitem{cho2014finding}
Minsu Cho, Jian Sun, Olivier Duchenne, and Jean Ponce.
\newblock Finding matches in a haystack: A max-pooling strategy for graph
  matching in the presence of outliers.
\newblock In {\em Proceedings of the IEEE Conference on Computer Vision and
  Pattern Recognition}, pages 2083--2090, 2014.

\bibitem{chui2003new}
Haili Chui and Anand Rangarajan.
\newblock A new point matching algorithm for non-rigid registration.
\newblock {\em Computer Vision and Image Understanding}, 89(2-3):114--141,
  2003.

\bibitem{chum2005matching}
Ondrej Chum and Jiri Matas.
\newblock Matching with prosac-progressive sample consensus.
\newblock In {\em 2005 IEEE computer society conference on computer vision and
  pattern recognition (CVPR'05)}, volume~1, pages 220--226. IEEE, 2005.

\bibitem{cross1998graph}
Andrew~DJ Cross and Edwin~R Hancock.
\newblock Graph matching with a dual-step em algorithm.
\newblock {\em IEEE transactions on pattern analysis and machine intelligence},
  20(11):1236--1253, 1998.

\bibitem{cuturi2013sinkhorn}
Marco Cuturi.
\newblock Sinkhorn distances: Lightspeed computation of optimal transport.
\newblock {\em Advances in neural information processing systems}, 26, 2013.

\bibitem{damelin2024near}
Steven~B Damelin.
\newblock {\em Near Extensions and Alignment of Data in Rn: Whitney extensions
  of near isometries, shortest paths, equidistribution, clustering and
  non-rigid alignment of data in Euclidean space}.
\newblock John Wiley \& Sons, 2024.

\bibitem{dang2020learning}
Zheng Dang, Fei Wang, and Mathieu Salzmann.
\newblock Learning 3d-3d correspondences for one-shot partial-to-partial
  registration.
\newblock {\em arXiv preprint arXiv:2006.04523}, 2020.

\bibitem{deng2022survey}
Bailin Deng, Yuxin Yao, Roberto~M Dyke, and Juyong Zhang.
\newblock A survey of non-rigid 3d registration.
\newblock In {\em Computer Graphics Forum}, volume~41, pages 559--589. Wiley
  Online Library, 2022.

\bibitem{deng2018ppfnet}
Haowen Deng, Tolga Birdal, and Slobodan Ilic.
\newblock Ppfnet: Global context aware local features for robust 3d point
  matching.
\newblock In {\em Proceedings of the IEEE conference on computer vision and
  pattern recognition}, pages 195--205, 2018.

\bibitem{deng20193d}
Haowen Deng, Tolga Birdal, and Slobodan Ilic.
\newblock 3d local features for direct pairwise registration.
\newblock In {\em Proceedings of the IEEE/CVF conference on computer vision and
  pattern recognition}, pages 3244--3253, 2019.

\bibitem{derpanis2010overview}
Konstantinos~G Derpanis.
\newblock Overview of the ransac algorithm.
\newblock {\em Image Rochester NY}, 4(1):2--3, 2010.

\bibitem{deshpande2019max}
Ishan Deshpande, Yuan-Ting Hu, Ruoyu Sun, Ayis Pyrros, Nasir Siddiqui, Sanmi
  Koyejo, Zhizhen Zhao, David Forsyth, and Alexander~G Schwing.
\newblock Max-sliced wasserstein distance and its use for gans.
\newblock In {\em Proceedings of the IEEE/CVF Conference on Computer Vision and
  Pattern Recognition}, pages 10648--10656, 2019.

\bibitem{diez2012hierarchical}
Yago Diez, Joan Mart{\'\i}, and Joaquim Salvi.
\newblock Hierarchical normal space sampling to speed up point cloud coarse
  matching.
\newblock {\em Pattern Recognition Letters}, 33(16):2127--2133, 2012.

\bibitem{domokos2011nonlinear}
Csaba Domokos, Jozsef Nemeth, and Zoltan Kato.
\newblock Nonlinear shape registration without correspondences.
\newblock {\em IEEE Transactions on pattern analysis and machine intelligence},
  34(5):943--958, 2011.

\bibitem{du2007extension}
Shaoyi Du, Nanning Zheng, Shihui Ying, Qubo You, and Yang Wu.
\newblock An extension of the icp algorithm considering scale factor.
\newblock In {\em 2007 IEEE International Conference on Image Processing},
  volume~5, pages V--193. IEEE, 2007.

\bibitem{duchon1977splines}
Jean Duchon.
\newblock Splines minimizing rotation-invariant semi-norms in sobolev spaces.
\newblock In {\em Constructive theory of functions of several variables}, pages
  85--100. Springer, 1977.

\bibitem{eisenberger2020deep}
Marvin Eisenberger, Aysim Toker, Laura Leal-Taix{\'e}, and Daniel Cremers.
\newblock Deep shells: Unsupervised shape correspondence with optimal
  transport.
\newblock {\em Advances in Neural information processing systems},
  33:10491--10502, 2020.

\bibitem{fatras2021unbalanced}
Kilian Fatras, Thibault S{\'e}journ{\'e}, R{\'e}mi Flamary, and Nicolas Courty.
\newblock Unbalanced minibatch optimal transport; applications to domain
  adaptation.
\newblock In {\em International Conference on Machine Learning}, pages
  3186--3197. PMLR, 2021.

\bibitem{feydy2017optimal}
Jean Feydy, Benjamin Charlier, Fran{\c{c}}ois-Xavier Vialard, and Gabriel
  Peyr{\'e}.
\newblock Optimal transport for diffeomorphic registration.
\newblock In {\em Medical Image Computing and Computer Assisted Intervention-
  MICCAI 2017: 20th International Conference, Quebec City, QC, Canada,
  September 11-13, 2017, Proceedings, Part I 20}, pages 291--299. Springer,
  2017.

\bibitem{figalli2010new}
Alessio Figalli and Nicola Gigli.
\newblock A new transportation distance between non-negative measures, with
  applications to gradients flows with dirichlet boundary conditions.
\newblock {\em Journal de math{\'e}matiques pures et appliqu{\'e}es},
  94(2):107--130, 2010.

\bibitem{figalli2021invitation}
Alessio Figalli and Federico Glaudo.
\newblock {\em An invitation to optimal transport, Wasserstein distances, and
  gradient flows}.
\newblock European Mathematical Society Publishing House, 2021.

\bibitem{fischler1981random}
Martin~A Fischler and Robert~C Bolles.
\newblock Random sample consensus: a paradigm for model fitting with
  applications to image analysis and automated cartography.
\newblock {\em Communications of the ACM}, 24(6):381--395, 1981.

\bibitem{flamary2021pot}
R{\'e}mi Flamary, Nicolas Courty, Alexandre Gramfort, Mokhtar~Z. Alaya,
  Aur{\'e}lie Boisbunon, Stanislas Chambon, Laetitia Chapel, Adrien Corenflos,
  Kilian Fatras, Nemo Fournier, L{\'e}o Gautheron, Nathalie~T.H. Gayraud,
  Hicham Janati, Alain Rakotomamonjy, Ievgen Redko, Antoine Rolet, Antony
  Schutz, Vivien Seguy, Danica~J. Sutherland, Romain Tavenard, Alexander Tong,
  and Titouan Vayer.
\newblock Pot: Python optimal transport.
\newblock {\em Journal of Machine Learning Research}, 22(78):1--8, 2021.

\bibitem{genevay2016stochastic}
Aude Genevay, Marco Cuturi, Gabriel Peyr{\'e}, and Francis Bach.
\newblock Stochastic optimization for large-scale optimal transport.
\newblock {\em Advances in neural information processing systems}, 29, 2016.

\bibitem{gold1998new}
Steven Gold, Anand Rangarajan, Chien-Ping Lu, Suguna Pappu, and Eric Mjolsness.
\newblock New algorithms for 2d and 3d point matching: pose estimation and
  correspondence.
\newblock {\em Pattern recognition}, 31(8):1019--1031, 1998.

\bibitem{grave2019unsupervised}
Edouard Grave, Armand Joulin, and Quentin Berthet.
\newblock Unsupervised alignment of embeddings with wasserstein procrustes.
\newblock In {\em The 22nd International Conference on Artificial Intelligence
  and Statistics}, pages 1880--1890. PMLR, 2019.

\bibitem{guittet2002extended}
Kevin Guittet.
\newblock {\em Extended Kantorovich norms: a tool for optimization}.
\newblock PhD thesis, INRIA, 2002.

\bibitem{haehnel2003extension}
Dirk Haehnel, Sebastian Thrun, and Wolfram Burgard.
\newblock An extension of the icp algorithm for modeling nonrigid objects with
  mobile robots.
\newblock In {\em IJCAI}, volume~3, pages 915--920, 2003.

\bibitem{hirose2020bayesian}
Osamu Hirose.
\newblock A bayesian formulation of coherent point drift.
\newblock {\em IEEE transactions on pattern analysis and machine intelligence},
  43(7):2269--2286, 2020.

\bibitem{hoppe1992surface}
Hugues Hoppe, Tony DeRose, Tom Duchamp, John McDonald, and Werner Stuetzle.
\newblock Surface reconstruction from unorganized points.
\newblock In {\em Proceedings of the 19th annual conference on computer
  graphics and interactive techniques}, pages 71--78, 1992.

\bibitem{horn1987closed}
Berthold~KP Horn.
\newblock Closed-form solution of absolute orientation using unit quaternions.
\newblock {\em Josa a}, 4(4):629--642, 1987.

\bibitem{horn1988closed}
Berthold~KP Horn, Hugh~M Hilden, and Shahriar Negahdaripour.
\newblock Closed-form solution of absolute orientation using orthonormal
  matrices.
\newblock {\em JOSA A}, 5(7):1127--1135, 1988.

\bibitem{huang2008non}
Qi-Xing Huang, Bart Adams, Martin Wicke, and Leonidas~J Guibas.
\newblock Non-rigid registration under isometric deformations.
\newblock In {\em Computer Graphics Forum}, volume~27, pages 1449--1457. Wiley
  Online Library, 2008.

\bibitem{huang2019automatic}
Ruikun Huang, Junli Zhao, Fuqing Duan, Xin Li, Celong Liu, Xiaodan Deng,
  Zhenkuan Pan, Zhongke Wu, and Mingquan Zhou.
\newblock Automatic craniofacial registration based on radial curves.
\newblock {\em Computers \& Graphics}, 82:264--274, 2019.

\bibitem{huang2021comprehensive}
Xiaoshui Huang, Guofeng Mei, Jian Zhang, and Rana Abbas.
\newblock A comprehensive survey on point cloud registration.
\newblock {\em arXiv preprint arXiv:2103.02690}, 2021.

\bibitem{ilci2020high}
Veli Ilci and Charles Toth.
\newblock High definition 3d map creation using gnss/imu/lidar sensor
  integration to support autonomous vehicle navigation.
\newblock {\em Sensors}, 20(3):899, 2020.

\bibitem{jian2005robust}
Bing Jian and Baba~C Vemuri.
\newblock A robust algorithm for point set registration using mixture of
  gaussians.
\newblock In {\em Tenth IEEE International Conference on Computer Vision
  (ICCV'05) Volume 1}, volume~2, pages 1246--1251. IEEE, 2005.

\bibitem{jin2021two}
Kun Jin, Chaoyue Liu, and Cathy Xia.
\newblock Two-sided wasserstein procrustes analysis.
\newblock In {\em IJCAI}, pages 3515--3521, 2021.

\bibitem{johnson1999registration}
Andrew~Edie Johnson and Sing~Bing Kang.
\newblock Registration and integration of textured 3d data.
\newblock {\em Image and vision computing}, 17(2):135--147, 1999.

\bibitem{joshi1995problem}
Anupam Joshi and C-H Lee.
\newblock On the problem of correspondence in range data and some inelastic
  uses for elastic nets.
\newblock {\em IEEE transactions on neural networks}, 6(3):716--723, 1995.

\bibitem{kantorovich1942translocation}
Leonid~V Kantorovich.
\newblock On the translocation of masses.
\newblock In {\em Dokl. Akad. Nauk. USSR (NS)}, volume~37, pages 199--201,
  1942.

\bibitem{kantorovich1948problem}
Leonid~V Kantorovich.
\newblock On a problem of monge.
\newblock In {\em CR (Doklady) Acad. Sci. URSS (NS)}, volume~3, pages 225--226,
  1948.

\bibitem{karmarkar1984new}
Narendra Karmarkar.
\newblock A new polynomial-time algorithm for linear programming.
\newblock In {\em Proceedings of the sixteenth annual ACM symposium on Theory
  of computing}, pages 302--311, 1984.

\bibitem{kolouri2019generalized}
Soheil Kolouri, Kimia Nadjahi, Umut Simsekli, Roland Badeau, and Gustavo Rohde.
\newblock Generalized sliced {Wasserstein} distances.
\newblock {\em Advances in Neural Information Processing Systems}, 32, 2019.

\bibitem{kolouri2015radon}
Soheil Kolouri, Se~Rim Park, and Gustavo~K Rohde.
\newblock The radon cumulative distribution transform and its application to
  image classification.
\newblock {\em IEEE transactions on image processing}, 25(2):920--934, 2015.

\bibitem{kolouri2016sliced}
Soheil Kolouri, Yang Zou, and Gustavo~K Rohde.
\newblock Sliced wasserstein kernels for probability distributions.
\newblock In {\em Proceedings of the IEEE Conference on Computer Vision and
  Pattern Recognition}, pages 5258--5267, 2016.

\bibitem{lecun2015deep}
Yann LeCun, Yoshua Bengio, and Geoffrey Hinton.
\newblock Deep learning.
\newblock {\em nature}, 521(7553):436--444, 2015.

\bibitem{lee2021generalized}
Wonjun Lee, Rongjie Lai, Wuchen Li, and Stanley Osher.
\newblock Generalized unnormalized optimal transport and its fast algorithms.
\newblock {\em Journal of Computational Physics}, 436:110041, 2021.

\bibitem{leonard2013survey}
Christian L{\'e}onard.
\newblock A survey of the schr$\backslash$" odinger problem and some of its
  connections with optimal transport.
\newblock {\em arXiv preprint arXiv:1308.0215}, 2013.

\bibitem{levi2014smooth}
Zohar Levi and Craig Gotsman.
\newblock Smooth rotation enhanced as-rigid-as-possible mesh animation.
\newblock {\em IEEE transactions on visualization and computer graphics},
  21(2):264--277, 2014.

\bibitem{li2008global}
Hao Li, Robert~W Sumner, and Mark Pauly.
\newblock Global correspondence optimization for non-rigid registration of
  depth scans.
\newblock In {\em Computer graphics forum}, volume~27, pages 1421--1430. Wiley
  Online Library, 2008.

\bibitem{li2023measure}
Shiying Li and Caroline Moosmueller.
\newblock Measure transfer via stochastic slicing and matching.
\newblock {\em arXiv preprint arXiv:2307.05705}, 2023.

\bibitem{liu2019flownet3d}
Xingyu Liu, Charles~R Qi, and Leonidas~J Guibas.
\newblock Flownet3d: Learning scene flow in 3d point clouds.
\newblock In {\em Proceedings of the IEEE/CVF Conference on Computer Vision and
  Pattern Recognition}, pages 529--537, 2019.

\bibitem{liutkus2019sliced}
Antoine Liutkus, Umut Simsekli, Szymon Majewski, Alain Durmus, and
  Fabian-Robert St{\"o}ter.
\newblock Sliced-wasserstein flows: Nonparametric generative modeling via
  optimal transport and diffusions.
\newblock In {\em International Conference on Machine Learning}, pages
  4104--4113. PMLR, 2019.

\bibitem{lu2019deepvcp}
Weixin Lu, Guowei Wan, Yao Zhou, Xiangyu Fu, Pengfei Yuan, and Shiyu Song.
\newblock Deepvcp: An end-to-end deep neural network for point cloud
  registration.
\newblock In {\em Proceedings of the IEEE/CVF International Conference on
  Computer Vision}, pages 12--21, 2019.

\bibitem{luo2003unified}
Bin Luo and Edwin~R Hancock.
\newblock A unified framework for alignment and correspondence.
\newblock {\em Computer Vision and Image Understanding}, 92(1):26--55, 2003.

\bibitem{ma2018nonrigid}
Jiayi Ma, Jia Wu, Ji Zhao, Junjun Jiang, Huabing Zhou, and Quan~Z Sheng.
\newblock Nonrigid point set registration with robust transformation learning
  under manifold regularization.
\newblock {\em IEEE transactions on neural networks and learning systems},
  30(12):3584--3597, 2018.

\bibitem{ma2015non}
Jiayi Ma, Ji Zhao, and Alan~L Yuille.
\newblock Non-rigid point set registration by preserving global and local
  structures.
\newblock {\em IEEE Transactions on image Processing}, 25(1):53--64, 2015.

\bibitem{mcneill2006probabilistic}
Graham McNeill and Sethu Vijayakumar.
\newblock A probabilistic approach to robust shape matching.
\newblock In {\em 2006 International Conference on Image Processing}, pages
  937--940. IEEE, 2006.

\bibitem{mei2021cotreg}
Guofeng Mei, Xiaoshui Huang, Litao Yu, Jian Zhang, and Mohammed Bennamoun.
\newblock Cotreg: Coupled optimal transport based point cloud registration.
\newblock {\em arXiv preprint arXiv:2112.14381}, 2021.

\bibitem{meinguet1979multivariate}
Jean Meinguet.
\newblock Multivariate interpolation at arbitrary points made simple.
\newblock {\em Zeitschrift f{\"u}r angewandte Mathematik und Physik ZAMP},
  30(2):292--304, 1979.

\bibitem{mendes2016icp}
Ellon Mendes, Pierrick Koch, and Simon Lacroix.
\newblock Icp-based pose-graph slam.
\newblock In {\em 2016 IEEE International Symposium on Safety, Security, and
  Rescue Robotics (SSRR)}, pages 195--200. IEEE, 2016.

\bibitem{monge1781memoire}
Gaspard Monge.
\newblock M{\'e}moire sur la th{\'e}orie des d{\'e}blais et des remblais.
\newblock {\em Mem. Math. Phys. Acad. Royale Sci.}, pages 666--704, 1781.

\bibitem{myronenko2010point}
Andriy Myronenko and Xubo Song.
\newblock Point set registration: Coherent point drift.
\newblock {\em IEEE transactions on pattern analysis and machine intelligence},
  32(12):2262--2275, 2010.

\bibitem{myronenko2006non}
Andriy Myronenko, Xubo Song, and Miguel Carreira-Perpinan.
\newblock Non-rigid point set registration: Coherent point drift.
\newblock {\em Advances in neural information processing systems}, 19, 2006.

\bibitem{nguyen2022revisiting}
Khai Nguyen and Nhat Ho.
\newblock Revisiting sliced wasserstein on images: From vectorization to
  convolution.
\newblock {\em Advances in Neural Information Processing Systems},
  35:17788--17801, 2022.

\bibitem{orlin1988faster}
James Orlin.
\newblock A faster strongly polynomial minimum cost flow algorithm.
\newblock In {\em Proceedings of the Twentieth annual ACM symposium on Theory
  of Computing}, pages 377--387, 1988.

\bibitem{pan2019clustermap}
Zhichen Pan, Haoyao Chen, Silin Li, and Yunhui Liu.
\newblock Clustermap building and relocalization in urban environments for
  unmanned vehicles.
\newblock {\em Sensors}, 19(19):4252, 2019.

\bibitem{pauly2003multi}
Mark Pauly, Richard Keiser, and Markus Gross.
\newblock Multi-scale feature extraction on point-sampled surfaces.
\newblock In {\em Computer graphics forum}, volume~22, pages 281--289. Wiley
  Online Library, 2003.

\bibitem{peyre2019computational}
Gabriel Peyr{\'e}, Marco Cuturi, et~al.
\newblock Computational optimal transport: With applications to data science.
\newblock {\em Foundations and Trends{\textregistered} in Machine Learning},
  11(5-6):355--607, 2019.

\bibitem{Piccoli2014Generalized}
Benedetto Piccoli and Francesco Rossi.
\newblock Generalized wasserstein distance and its application to transport
  equations with source.
\newblock {\em Archive for Rational Mechanics and Analysis}, 211(1):335--358,
  2014.

\bibitem{pomerleau2015review}
Fran{\c{c}}ois Pomerleau, Francis Colas, Roland Siegwart, et~al.
\newblock A review of point cloud registration algorithms for mobile robotics.
\newblock {\em Foundations and Trends{\textregistered} in Robotics},
  4(1):1--104, 2015.

\bibitem{poto2017laser}
Vivien Pot{\'o}, J{\'o}zsef~{\'A}rp{\'a}d Somogyi, Tam{\'a}s Lovas, and
  {\'A}rp{\'a}d Barsi.
\newblock Laser scanned point clouds to support autonomous vehicles.
\newblock {\em Transportation Research Procedia}, 27:531--537, 2017.

\bibitem{puy2020flot}
Gilles Puy, Alexandre Boulch, and Renaud Marlet.
\newblock Flot: Scene flow on point clouds guided by optimal transport.
\newblock In {\em Computer Vision--ECCV 2020: 16th European Conference,
  Glasgow, UK, August 23--28, 2020, Proceedings, Part XXVIII}, pages 527--544.
  Springer, 2020.

\bibitem{qi2017pointnet}
Charles~R Qi, Hao Su, Kaichun Mo, and Leonidas~J Guibas.
\newblock Pointnet: Deep learning on point sets for 3d classification and
  segmentation.
\newblock In {\em Proceedings of the IEEE conference on computer vision and
  pattern recognition}, pages 652--660, 2017.

\bibitem{qi2017pointnet++}
Charles~Ruizhongtai Qi, Li Yi, Hao Su, and Leonidas~J Guibas.
\newblock Pointnet++: Deep hierarchical feature learning on point sets in a
  metric space.
\newblock {\em Advances in neural information processing systems}, 30, 2017.

\bibitem{qin2022rigid}
Hongxing Qin, Yucheng Zhang, Zhentao Liu, and Baoquan Chen.
\newblock Rigid registration of point clouds based on partial optimal
  transport.
\newblock In {\em Computer Graphics Forum}, volume~41, pages 365--378. Wiley
  Online Library, 2022.

\bibitem{rabin2011wasserstein}
Julien Rabin, Gabriel Peyr{\'e}, Julie Delon, and Marc Bernot.
\newblock Wasserstein barycenter and its application to texture mixing.
\newblock In {\em International Conference on Scale Space and Variational
  Methods in Computer Vision}, pages 435--446. Springer, 2011.

\bibitem{ramirez2020novel}
Guillem Ram{\'\i}rez, Rumen Dangovski, Preslav Nakov, and Marin
  Solja{\v{c}}i{\'c}.
\newblock On a novel application of wasserstein-procrustes for unsupervised
  cross-lingual learning.
\newblock {\em arXiv preprint arXiv:2007.09456}, 2020.

\bibitem{rangarajan1997robust}
Anand Rangarajan, Haili Chui, Eric Mjolsness, Suguna Pappu, Lila Davachi,
  Patricia Goldman-Rakic, and James Duncan.
\newblock A robust point-matching algorithm for autoradiograph alignment.
\newblock {\em Medical image analysis}, 1(4):379--398, 1997.

\bibitem{rasoulian2012group}
Abtin Rasoulian, Robert Rohling, and Purang Abolmaesumi.
\newblock Group-wise registration of point sets for statistical shape models.
\newblock {\em IEEE transactions on medical imaging}, 31(11):2025--2034, 2012.

\bibitem{rusinkiewicz2001efficient}
Szymon Rusinkiewicz and Marc Levoy.
\newblock Efficient variants of the icp algorithm.
\newblock In {\em Proceedings third international conference on 3-D digital
  imaging and modeling}, pages 145--152. IEEE, 2001.

\bibitem{rusu20113d}
Radu~Bogdan Rusu and Steve Cousins.
\newblock 3d is here: Point cloud library (pcl).
\newblock In {\em 2011 IEEE international conference on robotics and
  automation}, pages 1--4. IEEE, 2011.

\bibitem{sarode2019pcrnet}
Vinit Sarode, Xueqian Li, Hunter Goforth, Yasuhiro Aoki, Rangaprasad~Arun
  Srivatsan, Simon Lucey, and Howie Choset.
\newblock Pcrnet: Point cloud registration network using pointnet encoding.
\newblock {\em arXiv preprint arXiv:1908.07906}, 2019.

\bibitem{sato2020fast}
Ryoma Sato, Marco Cuturi, Makoto Yamada, and Hisashi Kashima.
\newblock Fast and robust comparison of probability measures in heterogeneous
  spaces.
\newblock {\em arXiv preprint arXiv:2002.01615}, 2020.

\bibitem{scetbon2022low}
Meyer Scetbon and Marco Cuturi.
\newblock Low-rank optimal transport: Approximation, statistics and debiasing.
\newblock {\em Advances in Neural Information Processing Systems},
  35:6802--6814, 2022.

\bibitem{scetbon2021low}
Meyer Scetbon, Marco Cuturi, and Gabriel Peyr{\'e}.
\newblock Low-rank sinkhorn factorization.
\newblock In {\em International Conference on Machine Learning}, pages
  9344--9354. PMLR, 2021.

\bibitem{schonemann1966generalized}
Peter~H Sch{\"o}nemann.
\newblock A generalized solution of the orthogonal procrustes problem.
\newblock {\em Psychometrika}, 31(1):1--10, 1966.

\bibitem{schrodinger1931umkehrung}
Erwin Schr{\"o}dinger.
\newblock {\em {\"U}ber die umkehrung der naturgesetze}.
\newblock Verlag der Akademie der Wissenschaften in Kommission bei Walter De
  Gruyter u~…, 1931.

\bibitem{sejourne2023unbalanced}
Thibault S{\'e}journ{\'e}, Cl{\'e}ment Bonet, Kilian Fatras, Kimia Nadjahi, and
  Nicolas Courty.
\newblock Unbalanced optimal transport meets sliced-wasserstein.
\newblock {\em arXiv preprint arXiv:2306.07176}, 2023.

\bibitem{sharma2011topologically}
Avinash Sharma, Radu Horaud, Jan Cech, and Edmond Boyer.
\newblock Topologically-robust 3d shape matching based on diffusion geometry
  and seed growing.
\newblock In {\em CVPR 2011}, pages 2481--2488. IEEE, 2011.

\bibitem{shen2018mining}
Yiru Shen, Chen Feng, Yaoqing Yang, and Dong Tian.
\newblock Mining point cloud local structures by kernel correlation and graph
  pooling.
\newblock In {\em Proceedings of the IEEE conference on computer vision and
  pattern recognition}, pages 4548--4557, 2018.

\bibitem{shen2021accurate}
Zhengyang Shen, Jean Feydy, Peirong Liu, Ariel~H Curiale, Ruben San
  Jose~Estepar, Raul San Jose~Estepar, and Marc Niethammer.
\newblock Accurate point cloud registration with robust optimal transport.
\newblock {\em Advances in Neural Information Processing Systems},
  34:5373--5389, 2021.

\bibitem{sinkhorn1964relationship}
Richard Sinkhorn.
\newblock A relationship between arbitrary positive matrices and doubly
  stochastic matrices.
\newblock {\em The annals of mathematical statistics}, 35(2):876--879, 1964.

\bibitem{sinko20183d}
Martin Sinko, Patrik Kamencay, Robert Hudec, and Miroslav Benco.
\newblock 3d registration of the point cloud data using icp algorithm in
  medical image analysis.
\newblock In {\em 2018 ELEKTRO}, pages 1--6. IEEE, 2018.

\bibitem{tache2009magnebike}
Fabien T{\^a}che, Wolfgang Fischer, Gilles Caprari, Roland Siegwart, Roland
  Moser, and Francesco Mondada.
\newblock Magnebike: A magnetic wheeled robot with high mobility for inspecting
  complex-shaped structures.
\newblock {\em Journal of Field Robotics}, 26(5):453--476, 2009.

\bibitem{tevs2012animation}
Art Tevs, Alexander Berner, Michael Wand, Ivo Ihrke, Martin Bokeloh, Jens
  Kerber, and Hans-Peter Seidel.
\newblock Animation cartography—intrinsic reconstruction of shape and motion.
\newblock {\em ACM Transactions on Graphics (TOG)}, 31(2):1--15, 2012.

\bibitem{torr2000mlesac}
Philip~HS Torr and Andrew Zisserman.
\newblock Mlesac: A new robust estimator with application to estimating image
  geometry.
\newblock {\em Computer vision and image understanding}, 78(1):138--156, 2000.

\bibitem{tsin2004correlation}
Yanghai Tsin and Takeo Kanade.
\newblock A correlation-based approach to robust point set registration.
\newblock In {\em Computer Vision-ECCV 2004: 8th European Conference on
  Computer Vision, Prague, Czech Republic, May 11-14, 2004. Proceedings, Part
  III 8}, pages 558--569. Springer Berlin Heidelberg, 2004.

\bibitem{Villani2009Optimal}
Cedric Villani.
\newblock {\em Optimal transport: old and new}.
\newblock Springer, 2009.

\bibitem{villani2021topics}
C{\'e}dric Villani.
\newblock {\em Topics in optimal transportation}, volume~58.
\newblock American Mathematical Soc., 2021.

\bibitem{wahba1990spline}
Grace Wahba.
\newblock {\em Spline models for observational data}.
\newblock SIAM, 1990.

\bibitem{wang2019non}
Lingjing Wang, Jianchun Chen, Xiang Li, and Yi Fang.
\newblock Non-rigid point set registration networks.
\newblock {\em arXiv preprint arXiv:1904.01428}, 2019.

\bibitem{wang2019coherent}
Lingjing Wang, Xiang Li, Jianchun Chen, and Yi Fang.
\newblock Coherent point drift networks: Unsupervised learning of non-rigid
  point set registration.
\newblock {\em arXiv preprint arXiv:1906.03039}, 2019.

\bibitem{wang2019deep}
Yue Wang and Justin~M Solomon.
\newblock Deep closest point: Learning representations for point cloud
  registration.
\newblock In {\em Proceedings of the IEEE/CVF international conference on
  computer vision}, pages 3523--3532, 2019.

\bibitem{wang2022partial}
Zi-Ming Wang, Nan Xue, Ling Lei, and Gui-Song Xia.
\newblock Partial wasserstein adversarial network for non-rigid point set
  registration.
\newblock {\em arXiv preprint arXiv:2203.02227}, 2022.

\bibitem{wilson1969use}
Alan~Geoffrey Wilson.
\newblock The use of entropy maximising models, in the theory of trip
  distribution, mode split and route split.
\newblock {\em Journal of transport economics and policy}, pages 108--126,
  1969.

\bibitem{yang2011thin}
Jinzhong Yang.
\newblock The thin plate spline robust point matching (tps-rpm) algorithm: A
  revisit.
\newblock {\em Pattern Recognition Letters}, 32(7):910--918, 2011.

\bibitem{yang2022learning}
Zixin Yang, Richard Simon, and Cristian~A Linte.
\newblock Learning feature descriptors for pre-and intra-operative point cloud
  matching for laparoscopic liver registration.
\newblock {\em arXiv preprint arXiv:2211.03688}, 2022.

\bibitem{yew2020rpm}
Zi~Jian Yew and Gim~Hee Lee.
\newblock Rpm-net: Robust point matching using learned features.
\newblock In {\em Proceedings of the IEEE/CVF conference on computer vision and
  pattern recognition}, pages 11824--11833, 2020.

\bibitem{yuan20163d}
Chi Yuan, Xiaoqing Yu, and Ziyue Luo.
\newblock 3d point cloud matching based on principal component analysis and
  iterative closest point algorithm.
\newblock In {\em 2016 International Conference on Audio, Language and Image
  Processing (ICALIP)}, pages 404--408. IEEE, 2016.

\bibitem{yuan2020deepgmr}
Wentao Yuan, Benjamin Eckart, Kihwan Kim, Varun Jampani, Dieter Fox, and Jan
  Kautz.
\newblock Deepgmr: Learning latent gaussian mixture models for registration.
\newblock In {\em Computer Vision--ECCV 2020: 16th European Conference,
  Glasgow, UK, August 23--28, 2020, Proceedings, Part V 16}, pages 733--750.
  Springer, 2020.

\bibitem{yuille1988computational}
Alan~L Yuille and Norberto~M Grzywacz.
\newblock A computational theory for the perception of coherent visual motion.
\newblock {\em Nature}, 333(6168):71--74, 1988.

\bibitem{zeng20173dmatch}
Andy Zeng, Shuran Song, Matthias Nie{\ss}ner, Matthew Fisher, Jianxiong Xiao,
  and Thomas Funkhouser.
\newblock 3dmatch: Learning local geometric descriptors from rgb-d
  reconstructions.
\newblock In {\em Proceedings of the IEEE conference on computer vision and
  pattern recognition}, pages 1802--1811, 2017.

\bibitem{zhang2021fast}
Juyong Zhang, Yuxin Yao, and Bailin Deng.
\newblock Fast and robust iterative closest point.
\newblock {\em IEEE Transactions on Pattern Analysis and Machine Intelligence},
  2021.

\bibitem{zhang1994iterative}
Zhengyou Zhang.
\newblock Iterative point matching for registration of free-form curves and
  surfaces.
\newblock {\em International journal of computer vision}, 13(2):119--152, 1994.

\bibitem{zhang2020deep}
Zhiyuan Zhang, Yuchao Dai, and Jiadai Sun.
\newblock Deep learning based point cloud registration: an overview.
\newblock {\em Virtual Reality \& Intelligent Hardware}, 2(3):222--246, 2020.

\bibitem{zinsser2005point}
Timo Zin{\ss}er, Jochen Schmidt, and Heinrich Niemann.
\newblock Point set registration with integrated scale estimation.
\newblock In {\em International conference on pattern recognition and image
  processing}, pages 116--119, 2005.

\end{thebibliography}

\newpage 
\onecolumn

\appendices

\section{Notations}\label{sec:notations}

\noindent Below, we set the notations consistently used in the main text and appendix in this paper. 
\begin{itemize}
\item $\mathbb{R}_+=[0,\infty)$, set of all non-negative real numbers. 
\item $\mathbb{R}_{++}=(0,\infty)$, set of all positive real numbers. 
\item $\mathbb{R}^D$, the space of the point clouds. In this paper, without specification, we set $D=2$ or $D=3$.

\item Source point set: $$ X=\{x^n\in\mathbb{R}^D\}_{n=1}^N=[x^{1},\ldots , x^{N}]^T\in\mathbb{R}^{N\times D}\footnote{In this paper, we do not distinguish matrix and set.}$$ 

\item Target point set: $$ Y=\{y^m\in\mathbb{R}^D\}_{m=1}^M=[y^{1},\ldots,y^{M}]^T\in \mathbb{R}^{M\times D}$$
\item Control points: $$ C=\{c^k\in\mathbb{R}^D\}_{k=1}^K=[c^1,\ldots , c^K]^T \in \mathbb{R}^{K\times D}$$

\item $A[n,:]$, the $n^{th}$ row of matrix $A$.
\item $\mathcal{D}\subset [1:N]$, a set of indices. 

\item $A[\mathcal{D},:]$, the sub-matrix of $A$, 
formed by taking the rows corresponding to the indices in 
$\mathcal{D}$, 
i.e. if $n\in\mathcal{D}$, then $A[n,:]$ is in $A[\mathcal{D}]$.  
\item $I_D$, $D\times D$ identity matrix. 
\item $1_N$, $N\times 1$ vector with all 1 entries. 
\item $0_N$, $N\times 1$ vector with all 0 entries. 
\item $x_d$, the $d^{th}$ component of $x\in \mathbb{R}^D$.
\item $f:\mathbb{R}^D\mapsto \mathbb{R}^D$, is a deformation model that maps source point cloud to target point cloud. In general 
$$f(x)=h(x)+B^Tx+\beta$$
where $h(x)$ is the nonrigid part of the model, $B\in \mathbb{R}^{D\times D}$ form the affine part and $\beta\in \mathbb{R}^D$ is the translation parameter. 

On particular case of the above model is 
$$f(x)=h(x)+R^TSx+\beta.$$
In this formulation, 
$S=\text{diag}(s_1,\ldots s_D)\in\mathbb{R}^{D\times D}$ is a diagonal scaling matrix where each $s_i>0$. And $R\in \mathbb{R}^{D\times D}$ is a $D$-dimensional rotation matrix.\\
It is worth noting that the model presented in \cite{du2007extension} uses the formulation 
$RSx+\beta$. To maintain consistency between the RBF and TPS models, we opt for the $R^TSX+\beta$ formulation. Thus, the ICP step \eqref{eq: optimal R} is slightly different from the equation (17) in this reference. 

\item $f[d]$, $d^{th}$ component of $f$, where $d\in[1:D]$. 
\item $O_D$, set of all $D\times D$ orthornormal matrices. 
\item $\hat{Y}:=[\hat{y}^1,\ldots,\hat{y}^N]^T\in \mathbb{R}^{N\times D}$, the estimated target points, i.e. for each $n$, $\hat{y}^n=f(x^n)$ where $f$ is a selected model.

\item correspondence between source and target: 

In this paper, the phrase ``correspondence between source and target'' encompasses two distinct but interconnected concepts.

First, during the step of solving the OPT problem \eqref{eq: solve opt} or \eqref{eq: solve 1D OPT}, we have the deformed source point set denoted as $\{\hat y^n=f(x^n)\}_{n=1}^N$. In this context, we say the optimal transportation plan $\gamma$ is a (soft) correspondence between the (deformed) source points and the target points.

Second, after updating a subset of $\hat{y}$ points using either \eqref{eq: update Y_hat opt} or \eqref{eq: update y_hat sopt}, the set $\{\hat{y}^n\}_{n\in \mathcal{D}}$ is transformed to align with (a subset of) the target set $\{y_m\}_{m=1}^M$. In this transformed state, the point set $\{\hat{y}\}_{n\in\mathcal{D}}$ becomes a ``representation'' of the target points. As a result, we refer to the set of pairs $\{(x^n, \hat{y}^n)\}_{n \in \mathcal{D}}$ as a (partial) correspondence between the source and the (representation of) target.

\item $\tilde{f}:\mathbb{R}^D\to \mathbb{R}$, the  one dimensional model defined as 
$$\tilde{f}(x)=\tilde{h}(x)+b^Tx+\tilde\beta$$
where $\tilde{h}$ is the nonlinear deformation, $b\in \mathbb{R}^D$ and $\tilde{\beta}\in \mathbb{R}$. 
Thus point deformation model $f$ can be regarded as $(\tilde{f}_1,\ldots \tilde{f}_d)$, i.e., $f(x)=[\tilde{f}_1(x),\ldots,\tilde{f}_d(x)]^T$.
\item $\|\cdot\|$, $L_2$ norm in $\mathbb{R}^D$. 
\item $\tilde{Y}=\{\tilde{y}^n\in\mathbb{R}\}_{n=1}^{N}=[\tilde{y}^1,\ldots,\tilde{y}_n]^T$, target points for $1D$ interpolation problem 
\begin{align}  \min_{\tilde{f}}\sum_{n=1}^n\|\tilde{f}(x^n)-\tilde{y}^n\|^2+\text{reg}(\tilde{f})  \label{eq: 1D interporlation}
\end{align}
where $\text{reg}(\tilde{f})$ is a regularization term. In this paper, the above problem particularly becomes \eqref{eq: 1D RBF obj} and \eqref{eq: 1D TPS obj} when we discuss RBF model and TPS model respectively. 
\item $\phi(\cdot,\cdot),\phi_G(\cdot,\cdot),\phi_T(\cdot,\cdot)$: $\phi$ is a kernel function, in particular:  $$\phi_G(x,c;\sigma^2)=e^{-\frac{\|c-x\|^2}{\sigma^2}}$$ is the Gaussian kernel function with $\sigma$ denoting the kernel width; 
$$\phi_T(x,c; D)=\begin{cases}
(\|x-c\|^2)\ln(\|x-c\|) & \text{if } D=2 \nonumber \\ 
\|x-c\| & \text{if }D=3 \nonumber \\ 
\|x-c\|^{4-D} & \text{otherwise}
\end{cases}$$
is the Thin-Plate-Spline (TPS) kernel function.
\item $\Phi\in\mathbb{R}^{N\times K}$, $\Phi_{sub}=\Phi[\mathcal{D},:]$, where $\mathcal{D}\subset[1:N]$: Kernel matrices where $\Phi_{k,n}=\phi(x_n,c_k)$ and sub-matrix whose row is selected from $\Phi$ via indices in $\mathcal{D}$. 
\item $\alpha^1,\ldots, \alpha^K\in \mathbb{R}^{D}$: the parameters for the kernel mapping $$x\mapsto h(x):=\sum_{k=1}^K\alpha^k\phi(x,c^k)$$ in the point registration problem. 
\item $\tilde{\alpha}_1,\ldots, \tilde{\alpha}_K\in \mathbb{R}$: parameters of the kernel mapping $$\tilde{x}\mapsto \tilde{h}(x):=\sum_{n=1}^n\alpha^k\phi(x,\tilde{c}^k)$$ in the 1D corresponding kernel interpolation problem \eqref{eq: 1D interporlation}.


\item $\mathcal{M}_+(\mathbb{R}^D)$: the set of positive finite Radon measures defined on $\mathbb{R}^D$.
\item $\mathcal{P}(\mathbb{R}^D)$: the set of all probability measures defined on $\mathbb{R}^D$. 
\item $\mu,\nu\in\mathcal{M}_+(\mathbb{R}^D)$: two positive finite Radon measures.

\item $\pi_1,\pi_2:(\mathbb{R}^D)^2\to \mathbb{R}^D$: the canonical projection mapping,
$$\pi_1(\hat{y},y)=\hat{y}; \pi_2(\hat{y},y)=y.$$

\item $A$: A Borel set in space $\mathbb{R}^D$ or $(\mathbb{R}^{D})^2$. 
\item $f_\#\gamma$ where $\gamma\in \mathcal{M}_+((\mathbb{R}^D)^2)$, $f:(\mathbb{R}^D)^2\to\mathbb{R}^D$. The pushforward measure of $\gamma$ under $f$. In particular, for each Borel set $A\subseteq \mathbb{R}^D$, 
$$f_\#\gamma(A)=\gamma(f^{-1}(A)).$$
\item $|\mu|$: total mass of measure $\mu$. 
\item $\Pi(\mu,\nu):=\{\gamma\in\mathcal{P}((\mathbb{R}^D)^2): \pi_{1\#}\gamma=\mu,\pi_{2\#}\gamma=\nu\}$. 
 
\item $\Pi(\frac{1}{N}1_N,\frac{1}{M}1_M):=\{
\gamma\in \mathbb{R}_+^{N\times M}: \gamma \frac{1}{M}1_M=\frac{1}{N}1_N, \gamma^T \frac{1}{N}1_N=\frac{1}{M}1_M
\}$, where $1_N$ ($1_M$) is $N$-size ($M$-size) vector whose entries are $1$.  
\item $\mu\leq\nu$:   denotes that $\mu$ is dominated by $\nu$, i.e., for each Borel set $A\subseteq \mathbb{R}^D$, $\mu(A)\leq \nu(A)$.  
\item $\Pi_\leq(\mu,\nu):=\{\gamma\in\mathcal{M}_+((\mathbb{R}^{D})^2): \pi_{1\#}\gamma\leq \mu, \pi_{2\#}\gamma\leq\nu\}.$

\item Optimal transport problem between $\mu,\nu\in \mathcal{P}(\mathbb{R}^D)$: $$OT(\mu,\nu):=\inf_{\gamma\in\Pi(\mu,\nu)}\int c(\hat{y},y)d\gamma(\hat{y},y).$$
\item Empirical Optimal transport problem: 

$\mu=\frac{1}{N}\sum_{n=1}^N\delta_{\hat y^n}, \nu=\frac{1}{M}\sum_{m=1}^M\delta_{y^m}$, where $\delta_{\hat y^n}$ denotes the Dirac measure centered at $x^n$ (and similarly for $\delta_{y^m}$). 
$$\text{OT}(\mu,\nu):=\min_{\gamma\in \Pi(\frac{1}{N}1_N,\frac{1}{M}1_M)}\sum_{n=1}^N\sum_{m=1}^Mc(\hat{y}^n,y^m)\gamma_{n,m}$$
\item OT Barycentric projection of $\nu$ with respect to $\mu$: 
Given $\gamma$, an optimal solution for above empirical OT problem, the following empirical measure is called baricentric projection of $\nu$ (with respect to $\mu$): 
$$\hat\nu:=\sum_{n=1}^N \frac{1}{N}\delta_{(M\gamma Y)[n:]^T}.$$
In this formula, $M\gamma Y\in\mathbb{R}^{N\times D}$ and the $N$-points set $\{M\gamma Y[1]^T,\ldots M\gamma Y[N]^T\}$ can be regarded as a ``representation'' of target point set $Y=\{y^1,\ldots y^m\}$.

\item Optimal partial transport problem between $\mu,\nu\in \mathcal{M}_+(\mathbb{R}^D)$: $$OPT_{\lambda}(\mu,\nu):=\inf_{\gamma\in\Pi_\leq(\mu,\nu)}\int c(\hat{y},y)d\gamma(\hat{y},y)+\lambda(|\mu|+|\nu|-2|\gamma|).$$
where $\lambda\in \mathbb{R}_+$ is the penalty term for the creation and the destruction of mass. 
\item Empirical optimal partial transport: 
$\mu=\sum_{n=1}^N\delta_{\hat y^n}, \nu=\sum_{m=1}^M\delta_{y^m}$.
\begin{align}
  \text{OPT}_{\lambda}(\mu,\nu):
&=\min_{\gamma\in\Pi_\leq(1_N,1_M)}
\sum_{n=1}^N\sum_{m=1}^Mc(\hat{y}^n,y^m)\gamma_{n,m}+\lambda (N+M-2|\gamma|). \nonumber   
\end{align}

\item Primal optimal partial transport: 
\begin{align}
  \text{primal-OPT}(\mu,\nu;\zeta):=\inf_{\gamma\in\Pi_\leq(\mu,\nu)}&\int c(\hat{y},y)d\gamma(\hat{y},y) \nonumber \\ 
  &\text{s.t. }|\gamma|= \zeta. \nonumber
\end{align}
where $\zeta\in[0,\min\{|\mu|,|\nu|\}]$ is the mass constraint constant. 
\item Empirical primal optimal partial transport: 
\begin{align}
  \text{primal-OPT}(\mu,\nu;\zeta):=\inf_{\gamma\in\Pi_\leq(1_N,1_M)}&\sum c(\hat{y}^n,y^m)\gamma_{n,m}(\hat{y},y) \nonumber \\ 
  &\text{s.t. }|\gamma|= \zeta. \nonumber
\end{align}

\item $L$: The Monge mapping in the OT problem or the optimal partial bijection in empirical OPT problem. In particular, it means 
$$L:[1:N]\supset \text{Dom}(L)\to \text{Range}(L)\subset [1:M],$$
or denoted as 
$$L:[1:N]\hookrightarrow [1:M],$$
where $\text{Dom}(L),\text{Range}(L)$ is the domain and range of mapping $L$ respectively. In addition, $L$ is a bi-jection. We refer section \ref{sec:background} for the detailed introduction and explanation of Monge mapping and corresponding concepts.

\item $\mathbb{S}^{D-1}:=\{\theta\in \mathbb{R}^D: \|\theta\|^2=1\}$: the uniform sphere in $\mathbb{R}^D$.

\item $\theta_1,\ldots, \theta_{t'} \in \mathbb{S}^{D-1}$: the projection parameters we use for sliced-OT or sliced-OPT. 

\item $T\in\mathbb{N}$: maximum number of iterations for a point cloud registration method. 
\item $\bar{X}=[Q_1, Q_2]\begin{bmatrix}
    \mathcal{R}\\ 
    0_{(N-(D+1))\times N}
\end{bmatrix}$, QR decomposition of $\bar{X}$.

On the left hand side, $\bar{X}:=[1_N, X]$. 

On the right hand side, $Q_1\in \mathbb{R}^{N\times (D+1)},Q_2\in \mathbb{R}^{N\times (N-(D+1))}$, $\mathcal{R}\in \mathbb{R}^{(D+1)\times (D+1)}$. $[Q_1,Q_2]$ is an $N\times N$ orthonormal matrix, and $\mathcal{R}$ is upper triangular matrix. 

\item $\delta>0$: a constant used in the computation of primal-OPT problem. We refer Appendix \ref{sec:opt} for more details. 
\item $\xi>0$: the weight parameter of entropic regularization in the Sinkhorn OT (entropic OT) problem. 


\end{itemize}

\section{Primal form of optimal partial transport problem and its computation}\label{sec:opt}

Similar to the OPT problem \eqref{eq: OPT} as we introduced in section \ref{sec:background}, there is a one-by-one relation between the Primal form OPT \eqref{eq: OPT primal empirical} and the classical OT problem (see Proposition 1 \cite{chapel2020partial}).
In particular, 
let $\tilde{\mu}:=\mu+(\|\nu\|-\zeta)\delta_{\tilde{\infty}}, \tilde{\nu}:=\nu+(\|\mu\|-\zeta)\delta_{\tilde{\infty}}$, where $\tilde\infty$ is an auxiliary point. 
Define cost function $\tilde{c}$ with 
\begin{align}
\tilde{c}(\hat{y},y) =\begin{cases}
c(\hat{y}^n,y^m) & \text{if }\hat{y}=\hat{y}^n, y=y^m  \\ 
\lambda  &\text{if }\hat{y}=\hat{y}^n, y=\tilde{\infty} \text{ or }\hat{y}=\tilde{\infty}, y=y^m\\ 
2\lambda+\delta &\text{if }\hat{y}=y=\tilde\infty 
\end{cases} \label{eq: tilde c}
\end{align}
where $\lambda,\delta>0$.
We define OT problem  
$$\text{OT}
(\tilde{\mu},\tilde{\nu})=\min_{\tilde\gamma\in \Pi(\tilde\mu,\tilde\nu)}\sum_{n,m}\tilde{c}(\hat{y}^n,y^m)\tilde{\gamma}_{ij}$$
where $x_{N+1},y_{M+1}=\tilde\infty$.
We have 

\begin{proposition}\label{pro: opt and ot}
The following function is a bijection between $\Pi(\tilde\mu,\tilde\nu)$ and $\Pi_\leq (\mu,\nu)$
$$\Pi(\tilde\mu,\tilde\nu)\ni\tilde\gamma\mapsto \gamma[1:N,1:M]\in \Pi_\leq(\mu,\nu).$$
One transportation plan $\tilde\gamma$ is optimal for above OT problem iff $\tilde\gamma[1:N,1:M]$ is optimal for \eqref{eq: OPT primal empirical}    
\end{proposition}

 By the above relation between primal-OPT and OT, one can use the linear programming solver \cite{bonneel2011displacement} to solve the Primal form OPT problem, whose complexity is $\mathcal{O}((N+M)NM)$.

\section{Relation between TPS-RPM method and Entropic OT problem}\label{sec:Sinkhorn}
In this section, we discuss the relations between Entropic Optimal Transport (EOT) and TPS-RPM. Based on this relation, we propose an improved version of the TPS-RPM method.

\textbf{Entropic optimal transport}. The EOT problem, also known as the Sinkhorn distance, was initially studied in the Schrodinger problem \cite{schrodinger1931umkehrung,wilson1969use,leonard2013survey}. It gained popularity in the machine learning community through the influential paper\cite{cuturi2013sinkhorn}, which demonstrates the significant computational speed-up achievable using the Sinkhorn-Knopp algorithm.  In classical EOT, the problem is formulated between two probability measures, $\mu$ and $\nu$, both of which belong to the space $\mathcal{P}(\mathbb{R}^D)$, as follows:
\begin{align}
\text{EOT}_{\xi}(\mu,\nu):=\inf_{\gamma\in\Pi(\mu,\nu)}\int_{(\mathbb{R}^D)^2}c(\hat{y},y)d\gamma+\xi I(\gamma) \label{eq: entropic OT},
\end{align}
where $\xi\in(0,\infty)$ is a constant and 
\begin{align}
  I(\gamma):=\text{KL}(\gamma\parallel \mu\otimes\nu):=\int_{(\mathbb{R}^D)^2}\ln\left(\frac{d\gamma}{d\mu\otimes d\nu}\right) d\gamma \label{eq: mutual information}.  
\end{align}
The term $I(\gamma)$ represents the ``mutual information'', which is strictly convex with respect to $\gamma$. When we consider the (normalized) $\gamma$ as a joint distribution over the pair of random variables \((\hat{Y}_\mu, Y_\nu)\), \(I(\gamma)\) quantifies the amount of information about \(Y_\nu\) that is revealed when \(\hat{Y}_\mu\) is observed, or vice versa. 

In the context of empirical distributions, where \(\mu = \sum_{n=1}^{N} \frac{1}{N} \delta_{\hat{y}^n}\) and \(\nu = \sum_{m=1}^{M} \frac{1}{M} \delta_{y^m}\), the EOT problem under consideration becomes:

\begin{align}
EOT_\xi(\mu,\nu)&:=\inf_{\gamma\in\Pi(1_N,1_M)}\sum_{n,m}c(\hat{y}^n,y^m)\gamma_{n,m}+\xi \sum_{n,m} \ln (\gamma_{n,m})\gamma_{n,m}+\mathcal{C}_1 \nonumber \\ 
&\simeq\inf_{\gamma\in\Pi(1_N,1_M)}\sum_{n,m}c(\hat{y}^n,y^m)\gamma_{n,m}+\xi \sum_{n,m} \ln (\gamma_{n,m})\gamma_{n,m} \label{eq: EOT empirical}
\end{align}
where $\mathcal{C}_1'=\xi\sum_{n,m}\ln(NM)$ is a constant term. The EOT problem \eqref{eq: entropic OT} can be solved via the Sinkhorn-Knopp algorithm \cite{peyre2019computational}.  

\textbf{Entropic optimal partial transport}. The EOT problem can be naturally extended into the optimal partial transport setting \cite{chizat2018scaling}.  In particular, given $\mu,\nu\in\mathcal{M}_+(\Omega)$, the entropic OPT problem (EOPT) is defined as 
\begin{align}
\text{EOPT}_{\lambda, \xi}(\mu,\nu):=\inf_{\gamma\in\Pi_\leq(\mu,\nu)}\int_{(\mathbb{R}^D)^2}c(\hat{y},y) d\gamma(\hat{y},y)+\lambda(|\mu-\pi_1\gamma|+|\nu-\pi_2\gamma|)+\xi I(\gamma) \label{eq: entropic OPT}.
\end{align}
and in the empirical setting, it becomes
\begin{align}
\text{EOPT}_{\lambda, \xi}(\mu,\nu)&:=\min_{\gamma\in\Pi_\leq(1_N,1_M)}\sum_{n,m}c(\hat{y}^n,y^m)\gamma_{n,m}+\lambda(N+M-2|\gamma|)+\xi \sum_{n,m}\gamma_{n,m}\ln(\gamma_{n,m}) +\mathcal{C}_1\nonumber\\
&\simeq\min_{\gamma\in\Pi_\leq(1_N,1_M)}\sum_{n,m}c(\hat{y}^n,y^m)\gamma_{n,m}+\lambda(N+M-2|\gamma|)+\xi \sum_{n,m}\gamma_{n,m}\ln(\gamma_{n,m}) \nonumber\\ 
&=\min_{\tilde{\gamma}\in \tilde{\Pi}(1_N,1_M)}\sum_{n,m}c(\hat{y}^n,y^m)\tilde\gamma_{n,m}-2\lambda\sum_{n=1}^N\sum_{m=1}^M\tilde\gamma_{n,m}+\xi \sum_{n=1}^N\sum_{m=1}^M\tilde\gamma_{n,m}\ln(\tilde{\gamma}_{n,m})+\mathcal{C}_2 \label{eq: empirical entropic OPT 2}
\end{align}
where in the last line, 
$\tilde{\Pi}\subset \mathbb{R}_+^{(N+1)\times (M+1)}$ is defined as set of $\tilde\gamma$ such that:
\begin{align}
&\gamma 1_{M+1}[1:N]= 1_N, \gamma^T 1_{N+1}[1:M]=1_M \nonumber\\ 
&\gamma 1_{M+1}[N+1]= M, \gamma^T 1_{N+1}[M+1]=N  \label{eq: extra cond}.
\end{align}
$\mathcal{C}_2=\lambda(N+M)$ is a constant. 
This equation holds from the relation between the proposition \ref{pro: opt and ot}, we also refer \cite{caffarelli2010free,bai2022sliced} for more details.
Note, as an optimization problem, the constraint \eqref{eq: extra cond} can be released since this constraint only involves the last row and column of $\tilde\gamma$, i.e., $\tilde\gamma[N+1,:]$ and $\tilde\gamma[:, M+1]$, which is not contained in the objective function \eqref{eq: empirical entropic OPT 2}. Remove the constant $\mathcal{C}$, the problem \eqref{eq: empirical entropic OPT 2} becomes 
\begin{align}
\min_{\tilde\gamma}&\sum_{n=1}^N\sum_{m=1}^Mc(\hat{y}^n,y^m)\tilde{\gamma}_{ij}-2\lambda \sum_{n=1}^N\sum_{m=1}^M\tilde{\gamma}_{ij}\ln(\tilde{\gamma}_{ij})+\xi \sum_{n=1}^N\sum_{m=1}^M \tilde{\gamma}_{ij}\ln(\tilde{\gamma}_{ij}) \label{eq: empirical entropic OPT 3} \\ 
&\text{s.t. }\tilde{\gamma} 1_M[1:N]=1_N, \tilde{\gamma}^T1_{N}[1:M]=1_M. \nonumber  
\end{align}

\textbf{Understanding TPS-RPM from the point of view of EOPT}. 
In TPS-RPM \cite{chui2003new,yang2011thin} method, as we discussed before, the TPS model is defined as \eqref{eq: TPS model}: 
\begin{align}
f(x)=\sum_{n=1}^N\alpha^n\phi(x,x^n)+B^Tx+\beta \label{eq: TPS model 2}
\end{align}
and we aim to solve the following problem: 
\begin{align}
\min_{f,\hat\gamma}&\sum_{n=1}^N\sum_{m=1}^M\|y^m-f(x^n)\|^2\hat{\gamma}_{ij}+\lambda\|\nabla^2f\|^2-2\lambda \sum_{n=1}^N\sum_{m=1}^M\hat{\gamma}_{ij}+\xi \sum_{n=1}^N\sum_{m=1}^M\hat{\gamma}_{ij}\ln(\hat{\gamma}_{ij}) \label{eq: TPS-RPM obj}\\
&\text{s.t.}\hat{\gamma}\in\mathbb{R}_+^{(N+1)\times(M+1)},\hat\gamma 1_M[1:N]=1_N, \gamma^T1_N[1:M]=1_M. \nonumber
\end{align}
where $x^{N+1},y^{M+1}$ are auxiliary variables, called ``outlier center''. 

The method solves the problem described above in an iterative manner. Each iteration consists of two steps: Step 1 involves updating the correspondence $\hat{\gamma}$, and Step 2 involves updating $f$. Step 2 has been discussed in section \ref{sec:method} B and \ref{sec:method} D. Step 1 is essentially the entropic OPT problem \eqref{eq: empirical entropic OPT 3}. Indeed, when model $f$ is fixed, 
\begin{align}
&\min_{\hat\gamma}\sum_{n=1}^N\sum_{m=1}^M\|y^m-f(x^n)\|^2\hat{\gamma}_{ij}+\lambda\|\nabla^2f\|^2-2\lambda \sum_{n=1}^N\sum_{m=1}^M\hat{\gamma}_{ij}+\xi \sum_{n=1}^N\sum_{m=1}^M\hat{\gamma}_{ij}\ln(\hat{\gamma}_{ij}) \nonumber \\ 
\simeq&\min_{\hat\gamma}\sum_{n=1}^N\sum_{m=1}^M\|y^m-f(x^n)\|^2\hat{\gamma}_{ij}-2\lambda \sum_{n=1}^N\sum_{m=1}^M\hat{\gamma}_{ij}+\xi \sum_{n=1}^N\sum_{m=1}^M\hat{\gamma}_{ij}\ln(\hat{\gamma}_{ij})\nonumber 
\end{align}
One can either solve it via equations (3)--(7) in \cite{chui2003new}, or the Sinkhorn algorithm for entropic OPT problem \cite[Algorithm 1]{chizat2018scaling}. 

\textbf{An improved version of TPS-RPM}. 
By the above understanding, we consider the following entropic OPT problem: 
\begin{align}
&\min_{\gamma\in \Pi_\leq(\mu,\nu)}\int_{(\mathbb{R}^D)^2}c(\hat{y},y)d\gamma+\xi I(\gamma)\nonumber \\
&\quad\quad\quad\quad\text{s.t. } |\gamma|\ge \zeta. \label{eq: primal entropic OPT}
\end{align}
In the empirical setting, the above problem becomes (up to a constant)
\begin{align}
&\min_{\gamma\in\Pi_\leq(1_N,1_M)}\sum_{n,m}c(\hat{y}^n,y^m)\gamma_{n,m}+\xi\sum_{ij}\gamma_{n,m}\ln(\gamma_{n,m}) \label{eq: primal empirical entropic OPT}\\
&\quad\quad\quad\quad\quad\quad\text{s.t. }\sum_{n=1}\sum_{m=1}^M\gamma_{n,m}= \zeta. \nonumber 
\end{align}
Recall the relation between the primal-OPT and OT, we further simplify \eqref{eq: primal empirical entropic OPT} as 
\begin{align}
\min_{\hat\gamma}&\sum_{n=1}^N\sum_{m=1}^Mc(\hat{y}^n,y^m)\hat{\gamma}_{ij}+\lambda'(\sum_{n=1}^N\hat\gamma_{i,M+1}+\sum_{n=1}^M\hat\gamma_{N+1,j})+\delta\hat\gamma_{N+1,M+1}+\xi\sum_{n=1}^N\sum_{m=1}^M\hat{\gamma}_{ij}\ln(\hat{\gamma}_{ij}) \label{eq: primal empirical entropic OPT 2}\\
&\text{s.t. }
\hat\gamma\in\mathbb{R}_+^{(N+1)\times(M+1)}, \hat\gamma 1_M[1:N]=1_N,\hat\gamma^T 1_N[1:M]=1_M \nonumber\\  
&\hspace{1.5em}\hat\gamma1_M[N+1]=M-\zeta, \hat\gamma^T1_N[M+1]=N-\zeta. \nonumber 
\end{align}
where $\lambda',\delta>0$ are constants.
Note, un-rigorously speaking, the above problem is a balanced entropic OT problem and it can be solved by the Sinkhorn algorithm.  

Thus, we propose a new version of TPS-RPM method. Consider the following problem: 
\begin{align}
\min_{f,\gamma}&\sum_{n=1}^N\sum_{m=1}^M\|y^m-f(x^n)\|^2\gamma_{n,m}+\epsilon\|\nabla^2f\|^2+\xi \sum_{n=1}^N\sum_{m=1}^M\gamma_{n,m}\ln(\gamma_{n,m}) \label{eq: TPS-RPM obj2}\\
&\text{s.t. } \gamma\in\mathbb{R}_+^{N\times M},\hat\gamma 1_M\leq 1_N, \gamma^T1_N=1_M,\sum_{n=1}^N\sum_{m=1}^M\gamma_{n,m}=\zeta.\nonumber
\end{align}
In step of updating correspondence $\gamma$, it is equivalent to the entropic primal-OPT problem: 
\begin{align}
\min_\gamma &\sum_{n=1}^N\sum_{m=1}^M\|y^m-f(x^n)\|^2\gamma_{n,m}+\xi \sum_{n=1}^N\sum_{m=1}^M\ln(\gamma_{n,m})\gamma_{n,m} \nonumber\\
&\text{s.t. }\gamma\in\mathbb{R}^{N\times M}, \gamma 1_M\leq 1_N, \gamma^T 1_N\leq 1_M, |\gamma|= \zeta, \nonumber
\end{align}
One can either solve it via Sinkhorn algorithm for primal OPT problem  \cite{benamou2015iterative} or convert it into the classical entropic OT problem \eqref{eq: primal empirical entropic OPT 2}

\section{2D Experiment}\label{sec:2d_experiment}
In this section, we present an evaluation of various baseline methods, including CPD\cite{myronenko2010point}, the four non-rigid adaptations of Wasserstein Procrustes methods \cite{grave2019unsupervised}: OT-BBF, OT-TPS, SOT-RBF, and SOT-TPS, and the classical TPS-RPM \cite{chui2003new}, along with its modified variant TPS-RPM(new) discussed in the previous section, and the four novel methods proposed in Section \ref{sec:method} in 2D setting. The dataset for this experiment is the synthetic tropical fish, available at \url{https://github.com/siavashk/pycpd}. 
In this section, we present an evaluation of various baseline methods, including CPD\cite{myronenko2010point}, the four non-rigid adaptations of Wasserstein Procrustes methods \cite{grave2019unsupervised}: OT-BBF, OT-TPS, SOT-RBF, and SOT-TPS. Additionally, we include TPS-RPM \cite{chui2003new}, its modified variant TPS-RPM(new) discussed in the previous section, and the four novel methods proposed in Section \ref{sec:method}. Our dataset for this experiment is the synthetic tropical fish, available at \url{https://github.com/siavashk/pycpd}.

As in the 3D experiment, we introduce random noise to the data uniformly distributed over the dataset's support, $[-2,2]^2$. The noise level, denoted by $\eta$, represents the percentage of noisy data points relative to the clean dataset and varies within the set 
$\{0,10\%,20\%,30\%\}$. 
\begin{remark}

It's worth noting that the data used in this experiment do not fulfill the assumption outlined in equation \eqref{eq: RBF model} in Section \ref{sec:method}. To accentuate the differences between balanced and unbalanced optimal transport (OT) methods, noise is only added to point cloud $Y$. 
\end{remark}

\textbf{Accuracy}. 
\begin{figure}
    \centering
\includegraphics[width=1.0\textwidth]{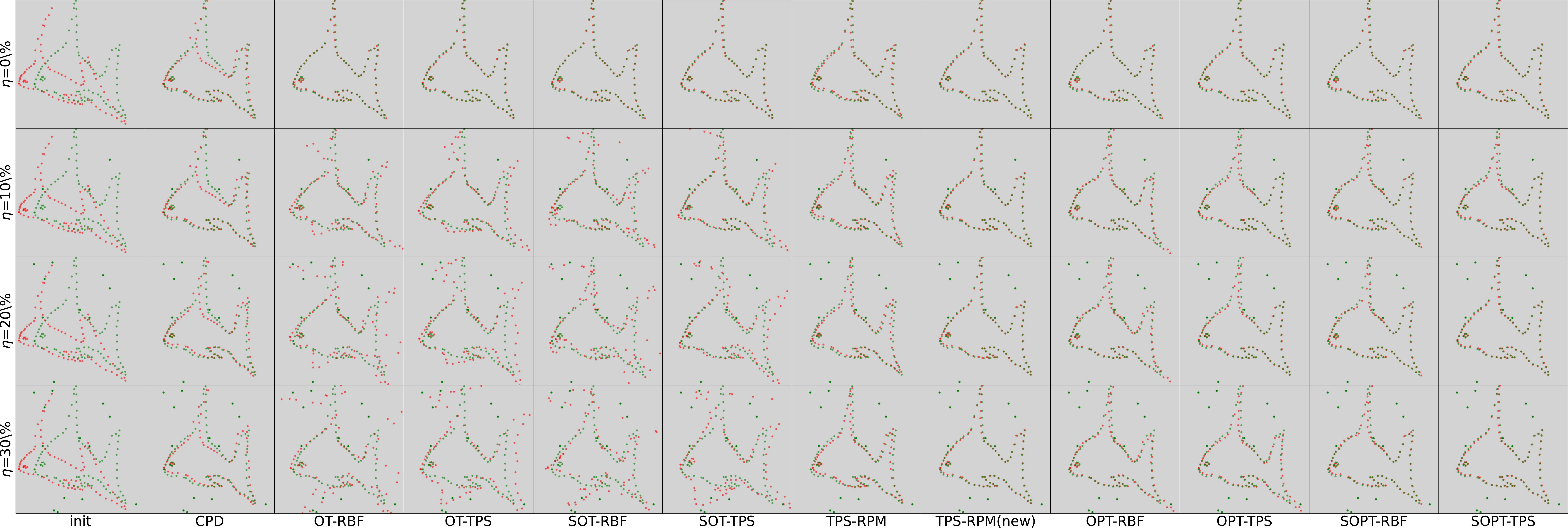}.
    \caption{We demonstrate the accuracy comparison of the 6 baseline methods: CPD\cite{myronenko2010point}, 4 nonrigid Wasserstein procrustes\cite{grave2019unsupervised,ramirez2020novel} methods, TPS-RPM \cite{chui2003new} and an variant introduced in last section, and our four partial OT based methods. $\eta$ is the noise level. The source and target point clouds are in red and green colors respectively.}
    \label{fig:2D-fish}
\end{figure}

See the figure \ref{fig:2D-fish} for the accuracy test. The parameter setting for each method is similar to the 3D experiments. We find that all methods are capable of successfully recovering the target point cloud when $\eta=0\%$. When $\eta$ increases to $10\%$, $20\%$, and $30\%$, the four balanced optimal transport (OT)-based methods—namely OT-RBF, OT-TPS, SOT-RBF, and SOT-TPS—result in relatively large errors.

Under conditions where the model constraint is not met, TPS-based methods (including both the old and new variants of TPS-RPM, as well as OPT-TPS and SOPT-TPS) generally exhibit superior performance. Furthermore, we notice that Sinkhorn-based methods (old and new TPS-RPM) and the sliced-OPT approach also perform well.

One plausible explanation for this could be that when the model constraint is not satisfied, the correspondence between the deformed source points $\{f(x^n)\}_{n=1}^N$ and the target points $\{y^m\}_{m=1}^M$, as determined by OPT, becomes limited reliable. In such cases, approaches incorporating soft correspondence, like the Sinkhorn method, or introducing randomness in the correspondence estimation step, like Sliced-OPT methods, result in more stable performance. 

Since the dataset is small (about $90$ to $100$), each method requires less than one second, we do not report the wall-clock time in this experiment. 
\begin{table}
\begin{center}
\textbf{Registration error of all methods in 2D fish experiment}\\
\vspace{1em}  
\begin{tabular}{|c| c | c | c |c|c|}
\hline
        &   &$\eta =0\%$ & $\eta =10\%$ & $\eta=20\%$ & $\eta=30\%$\\
\hline 
\multirow{7}{*}{\rotatebox[origin=c]{90}{\hspace{-0em} baselines}}
&CPD \cite{myronenko2010point}&0.171 &0.171&0.158 &0.158\\ 
&OT-RBF & 0.032 & 0.172 & 0.285 & 0.363 \\
&OT-TPS & 0.031 & 0.146 & 0.218  & 0.278 \\
&SOT-RBF& 0.032 & 0.206 &0.383  & 0.621 \\
&SOT-TPS& 0.031 & 0.213 &0.337  & 0.456\\ 
&TPS-RPM\cite{chui2003new} & 0.031 & 0.092 & 0.110 & 0.112 \\ 
&TPS-RPM(new)              & 0.032 & 0.035 & 0.037 & 0.038 \\ 
\hline 
\multirow{4}{*}{\rotatebox[origin=c]{90}{\hspace{-0em} ours}} 
&OPT-RBF&$0.032$  & $0.194$& $0.173$ & $0.192$\\
&OPT-TPS& $0.032$ & $0.175$ & $0.163$ & $0.145$\\
&SOPT-RBF& $0.031$ &$0.102$ & $0.138$ & $0.114$\\ 
&SOPT-TPS& $0.031$ & $0.032$ & $0.031$ & $0.033$\\
\hline 
\end{tabular}
\end{center}
\caption{This table displays the accuracy performance metrics for six baseline methods as well as our proposed methods. The variable $\eta$ is the noise level. For each pair of method and $\eta$, the corresponding cell contain the the error \eqref{eq: error}. }
\end{table}

\section{pseudo-code of OPT-TPS, SOPT-RBF, SOPT-TPS}\label{sec:code}
In this section, we present the pseudo-code for the algorithms described in Section \ref{sec:method}, Subsections B, C, and D. The input variables $X, Y, \zeta, T, \epsilon, C$ are elaborated upon in Subsection E of Section \ref{sec:method}. In Algorithms \ref{alg:opt-tps} and \ref{alg:sopt-TPS}, $\phi_T$ represents the TPS (Thin Plate Spline) kernel function. The corresponding parameter $D$ is denoted as `param', which can be manually specified.
Algorithms \ref{alg:sopt-RBF} and \ref{alg:sopt-TPS} make use of the variable $t$, representing the number of projections. In the first line of both methods, the variable $\lambda$ is initialized based on intuitive understanding, as outlined in \cite[Lemma 3.2]{bai2022sliced}. Generally, it is set to $K \|\text{mean}(Y) - \text{mean}(X)\|^2$, where $K \ge 2$.
\begin{algorithm}[H]
\caption{OPT-TPS}\label{alg:opt-tps}
\KwInput{$X,Y,\zeta,T,\phi_T,\text{param},\epsilon$}
\KwOutput{$B,\beta,\Phi,\alpha$}
\text{Initilize} $R,S,\beta,\alpha, B\gets SR$\\
\text{Initilize} $\Phi\gets \phi_T(X,X^T,\text{param})$\\
\For{$T' =1,2,\ldots T$}{
 \Comment{Step 1}
 $\hat{Y}\gets \Phi \alpha+XB+\beta^T1_{N}$\\
run lines 5-8 in Algorithm \ref{alg:opt-rbf}\\
\Comment{Step 2}
\If{condition for non-rigid is False}
{
$\hat{Y}'\gets \hat{Y}_{sub}-\Phi_{sub} \alpha$\\
By $(X_{sub},\hat{Y}')$, update $R,S,\beta$ via \eqref{eq: optimal R},\eqref{eq: optimal S},\eqref{eq: optimal beta}, $B\gets SR$ 
}
\Else{
By $(X,\Phi,\hat{Y})$, update $B,\beta,\alpha$ via \eqref{eq: TPS solution4}
}

}
\end{algorithm}

\begin{algorithm}[H]
\caption{SOPT-RBF}\label{alg:sopt-RBF}
\KwInput{$X,Y,\zeta,T,t,\phi,\text{param},\epsilon,C\gets X$}
\KwOutput{$R,S,\beta,\Phi,\alpha$}
\text{Initilize} $R,S,\beta,\alpha,\lambda$\\
\text{Initilize} $\Phi\gets \phi(X,C^T,\text{param})$\\
\For{$T' =1,2,\ldots T$}{
 \Comment{Step 1}
 $\hat{Y}\gets\Phi\alpha+XSR+\beta^T1_N$\\ 
 Sample $\{\theta_{t'}\}_{t'=1}^t\subset \mathbb{S}^{D-1}$\\ 
$\mathcal{D}\gets \emptyset$\\
\For{t'=1,2,\ldots t}{
$\hat{Y}_{\theta_{t'}}=\theta^T_{t'}Y,Y_{\theta_{t'}}=\theta^T_{t'}Y$\\ 
Compute optimal tansportation plan $L_{t'}$ for $\text{OPT}_{\lambda}(\hat{Y}_{\theta_{t'}},Y_{\theta_{t'}})$. \\ 
$\mathcal{D}_{t'}\gets\text{Dom}(L_{t'})$\\
update $\hat{y}^n,\forall n\in \mathcal{D}_{t'}$ via \eqref{eq: update y_hat sopt}\\ 
If $\# (\mathcal{D}_{t'})<\zeta$, increase $\lambda$; otherwise, decrease $\lambda$. \\
$\mathcal{D}\gets \mathcal{D}\cup \mathcal{D}_{t'}$
}
$(X_{sub},\hat{Y}_{sub},\Phi_{sub})=(X[\mathcal{D},:],\hat Y[\mathcal{D},:],\Phi[\mathcal{D},:])$\\
\Comment{Step 2}
run lines 9-13 in Algorithm \ref{alg:opt-rbf}.
}
\end{algorithm}
\begin{algorithm}[H]
\caption{SOPT-TPS}\label{alg:sopt-TPS}
\KwInput{$X,Y,\zeta,T,t,\phi_T,\text{param},\epsilon$}
\KwOutput{$B,\beta,\Phi,\alpha$}
\text{Initilize} $R,S,\beta,\alpha,\lambda, B\gets SR$\\
\text{Initilize} $\Phi\gets \phi_T(X,X^T,\text{param})$\\
\For{$T' =1,2,\ldots T$}{
 \Comment{Step 1}
$\hat{Y}\gets \Phi\alpha+XB+\beta^T1_N$\\
run lines 5-14 in Algorithm \ref{alg:sopt-RBF}\\
\Comment{Step 2}
run lines 6-10 in Algorithm \ref{alg:opt-tps}.
}
\end{algorithm}
\begin{remark}
For clarity, although $\phi_T$ is not actually an input parameter in algorithms \ref{alg:opt-tps} and \ref{alg:sopt-TPS} due to the use of a fixed TPS kernel, we include it as an input for the sake of consistency with other two RBF-based algorithms.
\end{remark}

\end{document}